\documentclass[12pt,letterpaper,english]{article}
\usepackage{microtype}
\usepackage{amsmath} 
\usepackage{amssymb}
\usepackage{amsthm}
\usepackage{mathrsfs}
\usepackage{caption}
\usepackage{fontawesome}
\usepackage[font=small,labelfont=bf]{caption}
\usepackage{subcaption}
\usepackage{setspace}
\usepackage{multirow}
\usepackage{rotating}
\usepackage{array}
\usepackage{multibib}
\usepackage{blindtext}
\usepackage{mathtools}
\usepackage{siunitx}
\usepackage{colortbl}
\usepackage{float}
\usepackage[flushleft]{threeparttable}
\usepackage{algorithm}
\usepackage{multirow, makecell}
\makeatletter
\renewcommand{\ALG@name}{Pseudo-code}
\makeatother
\usepackage{algorithmic}

\usepackage{xr}
\usepackage{color}

\DeclareMathAlphabet{\mathcalligra}{T1}{calligra}{m}{n}

\DeclareMathOperator{\E}{\mathbb{E}}

\usepackage[sectionbib,round]{natbib}
\DeclareMathOperator*{\argmax}{arg\,max}
\DeclareMathOperator*{\argmin}{arg\,min}
\usepackage{times}
\usepackage[small,compact]{titlesec} 
\usepackage{abstract}

\usepackage[hang,flushmargin]{footmisc}
\usepackage{fullpage} 
\usepackage{hyperref}
\usepackage[compact]{titlesec}
\usepackage[section]{placeins}
\usepackage{footnote}
\usepackage{epstopdf}
\usepackage{placeins}
\usepackage[toc,page]{appendix}
\usepackage{blkarray}
\usepackage{stackengine}
\usepackage{accents}
\usepackage{enumitem}
\usepackage[mathscr]{euscript}
\usepackage{booktabs}  
\usepackage{nicefrac}
\usepackage{longtable}

\usepackage[utf8]{inputenc}
\usepackage{longtable}
\usepackage{graphicx}
\usepackage{epstopdf}
\usepackage{bbm}
\usepackage{dsfont}
\usepackage{comment}
\usepackage{pgfplots}
\pgfplotsset{compat=1.6}
\newcommand{\squishlist}{
   \begin{list}{$\bullet$}
    { \setlength{\itemsep}{0pt} \setlength{\parsep}{1pt}
      \setlength{\topsep}{1pt} \setlength{\partopsep}{1pt}
      \setlength{\leftmargin}{1.5em} \setlength{\labelwidth}{1em}
      \setlength{\labelsep}{0.5em} } }

\newcommand{\squishlisttwo}{
   \begin{list}{$\bullet$}
    { \setlength{\itemsep}{0pt} \setlength{\parsep}{0pt}
      \setlength{\topsep}{0pt} \setlength{\partopsep}{0pt}
      \setlength{\leftmargin}{1em} \setlength{\labelwidth}{1.5em}
      \setlength{\labelsep}{0.5em} } }

\theoremstyle{definition}
\newtheorem{assum}{Assumption}

\theoremstyle{definition}
\newtheorem{defn}{Definition}

\newcommand{\squishend}{
    \end{list}  }

\newcommand{\hema}[1]{\noindent{\textcolor{green}{\{{\bf Hema:}  #1\}}}}

\begin{document}
\title{Design and Evaluation of Personalized Free Trials}

\author{
       Hema Yoganarasimhan \thanks{We are grateful to an anonymous firm for providing the data and to UW-Foster High-Performance Computing Lab for providing us with computing resources. We thank the participants of the 2018 Marketing Science conference and the Triennial Choice Symposium. Thanks are also due to seminar audiences at the Johns Hopkins University, University of California Berkeley, and University of Maryland, and the USC Marshall Webinar series for their feedback. Please address all correspondence to: hemay@uw.edu, ebar@uw.edu.}  \\
        University of Washington\\
        \and
       Ebrahim Barzegary\footnotemark[1]\\
        University of Washington\\
        \and
       Abhishek Pani\footnotemark[1] \\
        Bright Machines
 }
\thispagestyle{empty}

\maketitle

\newpage
\thispagestyle{empty}
\begin{abstract} 
Free trial promotions, where users are given a limited time to try the product for free, are a commonly used customer acquisition strategy in the Software as a Service (SaaS) industry. We examine how trial length affect users' responsiveness, and seek to quantify the gains from personalizing the length of the free trial promotions. Our data come from a large-scale field experiment conducted by a leading SaaS firm, where new users were randomly assigned to 7, 14, or 30 days of free trial. First, we show that the 7-day trial to all consumers is the best uniform policy, with a 5.59\% increase in subscriptions. Next, we develop a three-pronged framework for personalized policy design and evaluation. Using our framework, we develop seven personalized targeting policies based on linear regression, lasso, CART, random forest, XGBoost, causal tree, and causal forest, and evaluate their performances using the Inverse Propensity Score (IPS) estimator. We find that the personalized policy based on lasso performs the best, followed by the one based on XGBoost. In contrast, policies based on causal tree and causal forest perform poorly. We then link a method's effectiveness in designing policy with its ability to personalize the treatment sufficiently without over-fitting (i.e., capture spurious heterogeneity). Next, we segment consumers based on their optimal trial length and derive some substantive insights on the drivers of user behavior in this context. Finally, we show that policies designed to maximize short-run conversions also perform well on long-run outcomes such as consumer loyalty and profitability.

\vspace{5mm}

\end{abstract}

\newpage
\section{Introduction}
\label{sec:intro}
\setcounter{page}{1}

\subsection{SaaS Business Model and Free Trials}
One of the major trends in the software industry over the last few years has been the migration of software firms from the perpetual licensing business model to the ``Software as a Service'' (SaaS) model. In the SaaS model, the software is sold as a service, i.e., consumers can subscribe to the software based on monthly or annual contracts. Global revenues for the SaaS industry now exceed 200 billion USD \citep{gartner_2019}. This shift in the business model has fundamentally changed the marketing and promotional activities of software firms. At a high level, since the software is now a service, sales and trials play a more important role than traditional advertising and brand building activities. In particular, a new type of customer acquisition strategy that has become popular is free trial promotions, where new users are given a limited time to try the software for free. 

While free trials are now almost universal in the SaaS industry, there is no consensus on how long these trials should be. We observe trial lengths ranging anywhere from one week to three months in the industry; e.g., Microsoft 365 offers a 30 days free trial whereas Google's G Suite offers a 14 days free trial. There are pros and cons associated with both long and short trials, and the optimal trial length can depend on the mechanism through which the free trial operates. A short trial period lowers the firm's acquisition costs since the length of time that the firm supports a non-paying user is short. It is also less likely to suffer free-riding concerns. For example, if consumers need the software for a specific/defined task which will take only a few days (e.g., set up a website or edit a document/image), a long free trial allows them use the product without paying for it whereas a short trial period will force the customer to purchase it for the period of use. On the other hand, a long trial period can create stickiness/engagement and increases switching-back costs. For example, if a consumer has already created or stored content in the software, it is costly to move it to the local disk or alternative software. Longer trial periods can also enhance consumer learning by giving them more time to learn how to use the software (e.g., product features, functionalities).

While the above arguments make a global case for shorter/longer trial periods, the exact mechanism at work and the magnitude of its effect can be heterogeneous across consumers. For instance, it may be better to give longer free trials to beginners because they need more time to learn the software and convert, whereas students may have higher conversion rates with shorter free trials because it prevents free-riding. In principle, if there is significant heterogeneity in consumers' response to the length of free trials, then SaaS firms may benefit from adopting a personalized free trial policy. Formally, a personalized targeting policy is defined as a mapping from the space of users' pre-treatment attributes to the treatment space. In the free trial context, a personalized policy is one that assigns each consumer a different trial length depending on her/his observed demographics and skills. Intuitively, SaaS firms can benefit from personalization if they can assign each consumer the trial length that maximizes her/his expected responsiveness.

Indeed, SaaS free trials are particularly well-suited to personalization because of a few good reasons. First, software services have zero marginal costs and there are no direct cost implications of offering different trial lengths to different consumers. Second, it is easy to implement a personalized free trial policy at scale for digital services, unlike physical products. Finally, consumers are less likely to react adversely to receiving different free trial lengths  (unlike other marketing mix variables, e.g., prices). Thus, personalization of free trials promotions is not only feasible, but can also lead to substantial improvements in consumer acquisition rates and revenues in this industry.

\subsection{Large-Scale Field Experiment}
\label{ssec:expt}
In this paper, we use data from a large-scale free trial experiment conducted by a major SaaS firm to design and evaluate the effectiveness of personalized free trial policies. The firm sells a suite of related software products (e.g., Microsoft 365, Google G Suite, Adobe Creative Cloud) and is the leading player in its category, with close to monopolistic market power. At the time of this study, the firm used to give users a 30-day free trial for each of its software products, during which they had unlimited access to the product. 

To evaluate the effectiveness of different trial lengths, the firm conducted a field experiment involving 337,724 consumers from six major geographic markets -- Australia and New Zealand, France, Germany, Japan, United Kingdom, and United States of America. During the experiment period, new users who started a free trial for one of the firm's products were randomly assigned to one of 7, 14, or 30-day trial length conditions. These three trial lengths were chosen because they are the most commonly used ones in the industry and represent a vast majority of the SaaS free trials. The subscription and retention decisions of users in the experiment were monitored for two years. The firm also collected data on users' pre-treatment characteristics (e.g., skill level and job) and post-treatment retention, revenue, and product usage behavior to examine the heterogeneity in users' responses to trial lengths.

\subsection{Research Agenda and Challenges}
\label{ssec:agenda_challenges}
Our goal is to use the data from this field experiment to understand how the length of free trials affects customer acquisition in the SaaS industry. In particular, we are interested in four key questions related to trial lengths and personalization in the SaaS industry. First, what is the best uniform free trial policy, i.e., if we want to give all users the same trial length, what is the best option -- 7, 14, or 30 days? Second, is there heterogeneity in users' responsiveness to trial lengths? If yes, what percentage of users respond the best to 30 day trials and what are their characteristics (i.e., demographics/skills), what percentage respond the best to 14 day trials, and so on. Related to this, we would like to shed light on the role of different mechanisms at play here (e.g., learning, free-riding) and present some evidence on the extent to which they are heterogeneous across consumers. Third, what are gains from adopting the optimal personalized free trial policy compared to uniform policies. Finally, how do personalized policies designed to maximize short-run outcomes (i.e., customer acquisition) perform on long-run metrics such as consumer loyalty and revenue?

To answer these questions, we seek to develop a framework that managers and researchers can use to design and evaluate personalized targeting policies, and identify the optimal policy. We face a few key challenges in this task. First, from a theoretical perspective, searching for the optimal policy is a non-trivial task because the cardinality of the policy space can be quite large. (In our setting there are $3^{987,840}$ possible policies.) Second, from an empirical perspective, the covariate space of consumers' attributes is high dimensional in most applications.\footnote{In our application, we have $987,840$ possible sub-regions.} So we will not have sufficient data in each sub-region of the covariate space to learn how well a policy performs there. We therefore require methods that can automatically pool sub-regions that show similar responsiveness to the firm's actions and learn the optimal policy at the right level of granularity.

A series of recent papers at the intersection of machine learning and causal inference have been developing methods to address these challenges and obtain individual-level treatment effects, which can then be used to personalize treatment assignment \citep{Athey_Imbens_2016, Wager_Athey_2018}. Similarly, a series of papers in marketing have combined powerful predictive machine learning models with experimental (or quasi-experimental) data to develop personalized targeting policies \citep{Rafieian_Yoganarasimhan_2020, Rafieian_2019a, Rafieian_2019b, Simester_etal_2019a}. At a high level, all these papers share the common goal of personalizing marketing interventions in order to maximize some measure of reward that is important to the firm. However, the optimal policy that each of these papers/methods arrive at in a given empirical context can be very different because they all impose very different assumptions on the data generating process, are trained to optimize different objective functions, make different types of bias-variance trade-offs in the learning process, and have vastly different compute times. Thus far, we have little to no understanding of how these methods compare to each other when it comes to designing the optimal targeting policy. 

This brings us to the third challenge. In order to understand which of these methods produces the most effective personalized policy, we need to be able to evaluate the performance of each policy {\it offline} (without deploying it in the field). This is essential because deploying a policy in the field to estimate its effectiveness is costly in time and money. Moreover, given the size of the policy space, it is simply not feasible to test each policy in the field.

\subsection{Our Approach}
\label{ssec:our_approach}
In this paper, we provide a three-pronged framework for personalized policy design and evaluation that addresses these challenges. In the first component, we define the optimal policy design problem and present the solution concepts. As discussed earlier, an unstructured search for the optimal policy is challenging because the policy space can be very high dimensional. We circumvent this problem by adopting a two-step approach. In the first stage, we learn flexible functions of either outcomes or pairwise heterogeneous treatment effects. Then, in the second stage, we use the function(s) learned in the first stage to assign the optimal treatment for each user. Thus, we can design personalized policies if we have consistent estimates of either outcomes or  heterogeneous treatment effects.

In the second component, we design personalized policies based on five outcome estimators (linear regression, lasso, CART, random forest, and boosted regression trees) and two heterogeneous treatment effects estimators (causal tree and causal forest). All these methods impose different parametric or semi-parametric assumptions on users' responses or heterogeneous treatment effects. As such, while they should all give the same optimal policy when response functions are relatively simple and data are infinite, neither of these conditions is likely to be true in most applications.

The third component of our framework consists of offline policy evaluation. For this task, we use the Inverse Propensity Score (IPS) reward estimator, which has been used in the reinforcement learning and counterfactual policy evaluation literatures in computer science \citep{Horvitz_Thompson_1952, Dudik_etal_2011}. Intuitively, for {\it any} given policy, the estimator takes all the users who happened to receive the policy-prescribed treatment and scales them up by their propensity of receiving that treatment. This scaling gives us a pseudo-population that has received the policy being considered. The observed reward of this pseudo-population is, therefore, a consistent estimate of the reward that we can expect from the policy if we were to implement it. 

\subsection{Findings and Contribution}
\label{ssec:findings}

We apply this framework to data from the free trial experiment and present a set of substantive findings. At the outset, we find that the firm can do significantly better by simply assigning the 7-days trial to all consumers (which is the best uniform policy). In the test data, this leads to a 5.59\% gain in subscriptions, over the baseline of 30 days for all policy. In contrast, 14-days for all policy does not significantly increase subscriptions. This suggests that, simply shortening the free trial to 7 days will lead to a significant increase in subscription.

Next, we present a series of results comparing the performance of seven personalized targeting policies designed based on five outcome prediction estimators (linear regression, lasso, CART, random forest, and boosted regression trees or XGBoost) and two heterogeneous treatment effects estimators (causal tree and causal forest). First, we find that the treatment allocations vary significantly depending on the estimator used to design the policy. For example, the policy based on lasso prescribes the 7 day trial to 69\% of users whereas the policy based on causal forest prescribes the 7 day trial to 91\% of users. Second, we find that the personalized policy based on lasso performs the best, with a $6.81\%$ gain over the baseline of 30 days for all, followed by the policy based on XGBoost ($6.17\%$ improvement). However, policies based on other outcome estimators (e.g., random forests, regressions) perform poorly. Interestingly, policies based on the recently popular heterogeneous treatment effects estimators (causal tree and causal forest) also perform poorly. Causal tree is unable to personalize the policy at all. Causal forest personalizes policy by a small amount, but the gains from doing so are marginal. This an important finding since these methods are gaining traction in the marketing literature \citep{Guo_etal_2017, Fong_etal_2019}. However, it is not obvious that personalizing targeting interventions based on these methods is ideal or appropriate. Indeed, a key takeaway from our results is that managers may be better off simply going with the best uniform policy instead of investing resources in personalizing policies based on these methods.

We then adopt a descriptive approach to investigate the source of the difference in the performance of these policies. We find that methods like CART, causal tree, and casual forest perform poorly because they are unable to personalize the policy sufficiently. In contrast, methods like linear regression and random forest seem to infer excessive spurious heterogeneity, which is also problematic. Lasso and XGBoost lie somewhere in between these two groups -- these estimators are able to capture heterogeneity in users' responsiveness without over-fitting, and therefore perform the best when it comes personalized policy design. We then propose a new metric for quantifying the gains from implementing a specific policy compared to the baseline experiment, and demonstrate its usefulness in explaining policy performance.

Next, we focus on a substantive question of relevance to free trials in the SaaS industry: why do some users respond well to longer trials, while others respond better to shorter trials? To answer this question, we segment consumers into three groups based on their optimal trial length. We find that consumers for whom the 30-day trial is optimal are more likely to be {\it legacy users} (e.g.,  are more experienced and less likely to be students or hobbyists), consumers for whom the 14-day trial is optimal are more likely to be {\it learners} (e.g., have higher usage and subscription rate), and consumers for whom the 7-day trial is optimal are more likely to be {\it free-riders} (e.g., more likely to be beginners and students and have the lowest subscription rate). 

Finally, we find that the best-personalized policy (the one based on lasso) also performs well on long-term metrics, with a $7.96\%$ increase in subscription length and $11.61\%$ increase in revenues. Note that the magnitude of gains on these outcomes is different from that on subscriptions. We show that this is because trial length can affect long-run outcomes through multiple channels (in addition to conversion). Overall, our findings suggest that personalized targeting policies designed to maximize short-run conversions also perform well on long-run outcomes.

Our research makes two broad contributions to the literature. First, from a substantive perspective, we quantify the role of trial lengths in customer acquisition in the SaaS industry. We find that personalized free trials can significantly improve both short-run conversions and long-run profitability. Second, from a methodological perspective, we present a framework that managers and researchers can use to design and evaluate the effectiveness of personalized targeting strategies. While our findings on the relative performance of different estimators are specific to our context, the framework itself is quite general and can be used to personalize other marketing mix variables. 

The rest of the paper is organized as follows. In $\S$\ref{sec:related_lit} we discuss the related literature. In $\S$\ref{sec:app}, we describe the application setting and data, and we present a series of descriptive analysis in $\S$\ref{sec:descriptive}. Next, in $\S$\ref{sec:per_pol}, we describe the three components of our frameworks -- personalized policy design, estimators used to design policies, and policy evaluation. In $\S$\ref{sec:counter_policies}, we apply this framework to our data and discuss the performance of a series of uniform and personalized counterfactual policies. In $\S$\ref{sec:sub_finding}, we present some descriptives on consumer segmentation under the optimal personalized policy, and examine how customer loyalty and revenues evolve under this policy. We conclude in $\S$\ref{sec:conclusion}.

\section{Related Literature}
\label{sec:related_lit}

First, our paper relates to the theoretical and empirical research on policy design and evaluation that spans marketing, economics, and computer science. In an early theoretical paper, \citet{Manski_2004} presents a method that finds the optimal treatment for each observation by maximizing a regret function. More recent papers in this area include \cite{Kitagawa_Tetenov_2018} and \cite{Athey_Wager_2017}. However, the methods in these papers are constrained by the size of the policy space (which in our case is $3^{987,840}$). Therefore, we first estimate outcomes or heterogeneous treatment effects, and then use these estimates to develop and evaluate personalized policies.\footnote{There is another stream of literature within marketing that seeks to learn consumers preferences using active machine learning methods \citep{toubia_etal_2003, dzyabura_hauser_2011, dzyabura_hauser_2019}. In this case, the goal is preference discovery, not personalization and it mainly relies on conjoint experiments to achieve this task.}

Recent empirical papers in this area include \citet{Lefortier_etal_2016}, \citet{Swaminathan_etal_2017}, and \citet{Simester_etal_2019a}. In a particularly relevant paper, \citet{Simester_etal_2019b} investigate how data from field experiments can be used to design targeting policies for new customers or new regimes. They find that model-based methods (e.g., lasso) offer the best performance, though this advantage vanishes if the setting and/or consumers change significantly. Our paper differs from theirs both substantively and methodologically. Methodologically, our paper differs from theirs in three key ways. First, they evaluate effectiveness of different policies by running a second field experiment, whereas we use the offline IPS estimator. So our framework avoids the costs associated with a running a second field experiment. Second, unlike them, we also consider policies based on the newly proposed heterogeneous treatment effects estimators (e.g., causal forest). Third, we present descriptive approaches and metrics that are designed to explain the difference in the performance of the different policies. Finally, from a substantive perspective, we focus on free trials in the SaaS domain, whereas they study mail promotions for a retail firm. 

Our paper also adds to the small but growing marketing literature on applications of tree-based heterogeneous treatment effects estimators. \cite{Ascarza_2018} estimates the heterogeneous effect of retention interventions and shows that customers with the highest churn risk are not necessarily those who benefit most from retention interventions based on the method proposed by \citet{Guelman_etal_2015}. More recently, \cite{Guo_etal_2017} investigates the heterogeneous effect of information transparency on payments between pharmaceutical firms and physicians using causal forest \citep{Wager_Athey_2018}. Similarly, \citet{Fong_etal_2019} use the method to explore the heterogeneity in the returns to targeted promotions based on an individual's purchase history.  Our paper contributes to this literature by demonstrating that causal forest, and its predecessor causal tree, are unable to fully capture the heterogeneity in users' responses, and therefore are not very effective for personalizing targeting interventions.

Substantively, our paper adds to two important areas in digital marketing -- personalization of marketing actions and free trials. We discuss both in detail below. 

A recent stream of marketing papers has focused on the question of ``how to personalize marketing actions using machine learning methods in real-time'': website design \citep{hauser_etal_2009}, ranking of search engine results \citep{Yoganarasimhan_2020}, mobile ads \citep{Rafieian_Yoganarasimhan_2020}, the sequence of ads shown in a mobile in-app setting \citep{Rafieian_2019a, Rafieian_2019b}, and promotions \citep{Hitsch_Misra_2018}. We contribute to this research by presenting a framework for designing and evaluating personalized targeting policies and a case study on personalized free trial promotions.

Finally, our work relates to research that examines the effectiveness of free trials. Analytical papers in this area have explored mechanisms by which free trials can be effective. They have proposed a multitude of (often conflicting) explanations such as switching costs, software complexity, network effects, quality signaling, and consumer learning \citep{Cheng_Liu_2012, Dey_etal_2013, Wang_Ozkan_2018}. However, there are only two empirical papers that focus on free trials. \citet{Foubert_Gijsbrechts_2016} build a model of consumer learning and show that while free trials can enhance adoption, ill-timed free trials can also suppress adoption. Using a bayesian learning approach, \citet{Sunada_2018} compares the profitability of different free trial configurations. However, neither of these papers explore the question of how to optimize the length of free trials because they have no variation in the length of the free trials offered in their data. In contrast, we use data from a large-scale field experiment with exogenous variation in the length of free trials to identify the optimal trial length for each user.

\section{Setting and Data}
\label{sec:app}
In this section, we describe our application setting and data, and present some simple descriptive analyses that support the idea that personalizing free trial lengths can improve firm-level outcomes.

\subsection{Setting}
\label{ssec:setting}
Our data come from a major SaaS firm that sells a suite of software products. The suite includes a set of related software products (similar to Excel, Word, PowerPoint in Microsoft's MS-Office). The firm is the leading player in its category, with close to monopolistic market power. Users can either subscribe to single-product plans that allow them access to one software product or to bundled plans that allow them to use several products at the same time. Bundles are designed to target specific segments of consumers and consist of a set of complementary products. The prices of the plans vary significantly and depend on the bundle, the type of subscription (regular or educational), and the length of commitment (monthly or annual). Standard subscriptions run from $\$30$ to $\$140$ per month depending on the products in the bundle and come with a monthly renewal option. (To preserve the firm's anonymity, we have multiplied all the dollar values in the paper by a constant number.) If the user is willing to commit to an annual subscription, they receive over $30\%$ discount in price. However, users in annual contracts have to pay a sizable penalty to unsubscribe before the end of their commitment. The firm also offers educational licenses at a discounted rate to students and educational institutions, and these constitute $20.8\%$ of the subscriptions in our data. 

\subsection{Field Experiment}
\label{ssec:field_expt}

At the time of this study, the firm used to give users a 30-day free trial for each of its software products, during which they had unlimited access to the product.\footnote{This free trial is at the software product level, i.e., users start a separate trial for each software product, and their trial for a given product expires 30 days from the point at which they started the free trial for it.} In order to access the product after the trial period, users need a subscription to a plan or bundle that includes that product.

To evaluate the effectiveness of different trial lengths, the firm conducted a large-scale field experiment that ran from December $1^{st} 2015$ to January $6^{th} 2016$ and spanned six major geographic markets -- Australia and New Zealand, France, Germany, Japan, United Kingdom, and United States of America. During the experiment period, users who started a free trial for any of the firm's four most popular products were randomly assigned to one of 7, 14 or 30 days free trial length buckets. The assignment was at user level, i.e., once a user was assigned to a treatment (trial length), her/his trial length for the other three popular products was also set at the same length. The length of the free trial for other products during this period remained unchanged at 30 days. The summary statistics for the treatment assignment and subscriptions are shown in Table \ref{table:treat_summary}.

\begin{table}[t]
\centering
\footnotesize
\begin{tabular}{l|rrrr}
\toprule
{} &   \textbf{7 Days Trial} &  \textbf{14 Days Trial} &  \textbf{30 Days Trial} & \textbf{Total}\\
\hline\hline
Number of observations (N) & 51,040 & 51,017 &   235,667 &    337,724 \\
Percent of total observations               &  15.11 &              15.11 &              69.78 & 100 \\
\midrule
Number of subscriptions & 7,835    &  7,635 &   34,564 &  50,034 \\
Percent of total subscriptions & 15.66               &  15.26 &              69.08 &              100 \\
Subscription rate within group (in \%) & 15.36 & 14.96 & 14.67 & 14.81 \\
\bottomrule
\end{tabular}
\caption{\label{table:treat_summary}  Summary statistics of treatment assignment and subscription rates.}
\end{table}

The experiment was carefully designed and implemented to rule out the possibility of self-selection into treatments, a common problem in field experiments. In our setting, if users can see which treatment (or free trial length) they are assigned to prior to starting their trial, then users who find their treatment undesirable may choose to not start the trial. In that case, the observed sample of users in each treatment condition would no longer be random, and this in turn would bias the estimated treatment effects. Moreover, since the experimenter cannot obtain data on those who choose to not to start their free trials, there is no way to address this problem econometrically.

To avoid these types of self-selection problems, the firm designed the experiment so that users were informed of their trial-length only after starting their trial. In order to try a software product, users had to take the following steps: (1) sign up with the firm by creating an ID, (2) download an app manager that manages the download and installation of all the firm's products, and (3) click on an embedded \textit{start trial} button to start the trial for a given product. Only at this point in time, they are shown the length of their free trial as the time left before their trial expires (e.g., ``Your free trial expires in 7 days''). While users can simply quit or choose to not use the product at this point, their identities and actions are nevertheless captured in our data and incorporated in our analysis. 

In sum, the design satisfies the two main conditions necessary for the experiment to be deemed ``clean'' -- (1) unconfoundedness and (2) compliance \citep{Mutz_etal_2019}. 

\subsection{Data}
\label{ssec:data}
We have data on 337,724 users, who started a free trial for at least one of the four major products included in the field experiment. For each user $i$, we observe the following information -- (1) Treatment assignment ($W_i$), (2) Pre-treatment demographic data ($X_i$), and (3) Post-treatment behavioral data ($Z_i$). The treatment assignment variable denotes the trial length that the user was assigned to -- 7, 14, or 30 days. The variables under the latter two are described in detail below.
    
\subsubsection{Pre-treatment demographic data}
\label{sssec:pre_data}
  \begin{table}[t]
\centering
\footnotesize{
\begin{tabular}{l|c|cccc} 
\toprule
\multirow{2}{*}{\textbf{Variable} } & \multicolumn{1}{c|}{\textbf{Number of} } & \multicolumn{4}{c}{\textbf{Share of top sub-categories} }    \\ 
                                                \;\;\, & \textbf{sub-categories} & \textbf{ $1^{st}$} &  \textbf{$2^{nd}$} & \textbf{$3^{rd}$} &
                    \textbf{$4^{th}$}\\ \hline\hline
Geographic region &  6 &  55.02\% &  13.66\% &   9.12\% &   8.83\% \\
Operating system & 8 &  28.97\% &   21.4\% &  14.04\% &  13.98\% \\
Signup channel & 42 &  81.56\% &   8.14\% &   3.47\% &   0.81\% \\
Job & 14 &   28.20\% &   21.90\% &  20.34\% &   8.46\% \\
Skill & 5 &  69.05\% &  12.75\% &  10.77\% &   7.38\% \\
Business segment & 7 &  35.41\% &  32.74\% &   18.40\% &   7.81\% \\
\bottomrule
\end{tabular}
\caption{Summary statistics for the pre-treatment categorical variables.\label{table:Xsummary}}}
\end{table}
  
  \begin{enumerate}
  \item Geographic region: The geographic region/country that the user belongs to (one of the six described in $\S$\ref{ssec:setting}). It is automatically inferred from the user's IP address. 
  \item Operating system: The OS installed on the user's computer. It can take eight possible values, e.g., Windows 7, Mac OS Yosemite. It is inferred by the firm based on the compatibility of the products downloaded with the user's OS.
  \item Sign-up channel: The channel through which users came to sign-up for the free trial. In total, there are 42 possible sign-up channels, e.g., from the legacy version of the software, from the firm's website, through third-parties, and so on. 
  \item Skill: A self-reported measure of the user's fluency in using the firm's software suite. This can take four possible values -- beginner, intermediate, experienced, and mixed. 
  \item Job: The user's job-title (self-reported). The firm gives users 13 job-titles to pick from, e.g., student, business professional, hobbyist.
  \item Business segment: The self-reported business segment that the user belongs to. Users can choose from six options here, e.g., educational institution, individual, enterprise.
  \end{enumerate}
  
Note that the last three variables are self-reported though not open-ended, i.e., the firm gives users a list and requests them to pick one option from that list.  However, users may choose not to report these values, in which case, the missing values are recorded as ``unknown''. We treat this as an additional category for each of these three variables in our analysis.\footnote{Only a small fraction of people chooses to not report these data. For example, the percentage of users with ``unknown" Skill and Job is 7.4\% and 21.9\%, respectively.} The list of all the six pre-treatment variables and their summary statistics are shown in Table \ref{table:Xsummary}.
  
\subsubsection{Post-treatment behavioral data}
\label{sssec:post_data}
For all the users in our data, we observe all product download and usage data for the duration of their trial period. Further, we observe their subscription and renewal decisions for approximately 24 months (from December 2015 till November 2017). Using these data, we extract the following post-treatment behavioral information:
  \begin{enumerate}
  \item Subscription information: We have data on whether a user subscribes or not, and the date and type of subscription (product or bundle of products) if she does subscribe.
  
    \item Subscription length: Number of months that the user is a subscriber of one or more products/bundles during the 24-month observation period. If a user does not subscribe to any of the firm's products during the observation period, then this number is zero by default. If a user unsubscribes for a period of time and then comes back, her subscription length is the total number of months that she was a paying customer of the firm.\footnote{If a user subscribes to two or more plans, we aggregate the length of subscription all plans and report the total. So the subscription length can be greater than 24 months for such users.}
    
  \item Revenue: The total revenue (in scaled dollars) generated by the user over the 2-year observation period. This is a function of the user's subscription date, the products and/or bundles that she subscribes to, and her subscription length. 
  
  \item Products downloaded: The date and time-stamp of each product downloaded by the user.
  
  \item Usage information: Each product in the software suite has thousands of functionalities. Functionalities can be thought of as micro-tasks and are defined at the click and key-stroke level; e.g., save a file, click undo, and create a table. The firm captures all the usage data for users during their free trial and stores two key variables associated with usage:
  \begin{enumerate}
  \item Total usage: Total count of the functionalities used by the user during her trial period. 
  \item Distinct usage: Number of unique functionalities used by the user during her trial period.
  \end{enumerate}
  \end{enumerate}
  
The summary statistics of these post-treatment variables are presented in Table \ref{table:post_summary}. Both subscription length and revenue are shown for: (a) all users and (b) the subset of users who subscribed. There are a couple of points to note regarding the data on subscription length and revenues:
\squishlist
\item The minimum subscription length observed in the data for subscribers is zero (for 58 users). These are users who immediately (within one month) unsubscribed after subscribing, in which case the firm returns their money and records their subscription length and revenue as zero. 
\item We do not have access to the revenue data for team subscriptions and government subscriptions (which constitute a total of 3501 subscriptions). Hence, the number of observations used to calculate the summary statistics for revenue for subscribers is lower.
\squishend
The usage data are also missing (at random) for a subset of users and we report the summary statistics for non-missing observations.

\begin{table}[t]
\centering
\footnotesize
\begin{tabular}{l|cccccccc}
\toprule
\multirow{2}{*}{Variable}  & \multirow{2}{*}{\textbf{Mean}} &  \textbf{Standard} & \multirow{2}{*}{\textbf{Min}}& \multirow{2}{*}{\textbf{25\%}}& \multirow{2}{*}{\textbf{50\%}}&\multirow{2}{*}{\textbf{75\%}}& \multirow{2}{*}{\textbf{Max}} & \textbf{Number of}\\
& & \textbf{Deviation} & & & & & &\textbf{Observations} \\ 
\hline\hline
Subscription & 0.148 & 0.355 & 0 & 0 & 0 & 0 & 1 & 337,724\\
Subscription length (all) & 2.21 & 6.35 & 0 & 0 & 0 & 0& 108 & 337,724 \\
Subscription length (subscribers) & 16.02 & 8.43 & 0 & 10 & 17 & 22& 108 & 50,034\\
Revenue (all) & 79.13 & 285 & 0 & 0 & 0 & 0 & 20,208 & 334,223 \\
Revenue (subscribers) & 568 & 552 & 0 & 242 & 420 & 666 & 20,208  & 46,533\\
No. of products downloaded  & 1.15 & 0.40 & 1 & 1 & 1 & 1 & 4 & 337,724\\
Total usage  & 2,226 & 8,071 & 5 & 133 & 454 & 1,556 & 543,315 & 302,646\\
Distinct usage & 102.41 & 96.61 & 4 & 48 & 75 & 118 & 4,033 & 302,640\\
\bottomrule
\end{tabular}
\caption{\label{table:post_summary}  Summary statistics of post-treatment variables.}
\end{table}

\subsection{Training and Test Datasets}
\label{ssec:datasets}

\begin{table}[t]
\centering
\footnotesize
\begin{tabular}{l|cc|c}
\toprule
\textbf{Trial Length} &   \textbf{Training Dataset} &  \textbf{Test Dataset} & \textbf{Total}\\
\hline\hline
\textbf{7 days} & 35,743 & 15,274 & 51,017 \\
\textbf{14 days} & 35,901 & 15,139 & 51,040 \\
\textbf{30 days} & 165,056 & 70,611 & 235,667\\
\midrule
\textbf{Total} & 236,700 & 101,024 & 337,724\\
\bottomrule
\end{tabular}
\caption{\label{table:data_distribution} The number of observations for each trial length in each dataset.}
\end{table}

To implement and test any policy that we design, we first need to partition the data into two independent samples -- training data and test data. These are defined as follows:
\squishlist
\item {\bf Training Data:} This is the data that is used for both learning the model parameters as well as model selection (or hyper-parameter optimization through cross-validation). 
\item {\bf Test Data:} This is a hold-out data on which we can evaluate the performance of the policies designed based on the models built on training data. 
\squishend
We use 70\% of the data for training (and validation) and 30\% for test. See Table \ref{table:data_distribution} for a detailed breakdown of how the data are split across the two data-sets. Note that while the joint distributions of the variables in the two samples should be the same theoretically, there will be some minor differences between the two data-sets due to the randomness in splitting in a finite sample. It is important to keep this in mind when comparing results {\it across} the two data-sets.

\section{Descriptive Analysis}
\label{sec:descriptive}
We now present some model-free evidence on the effect of trial length on subscriptions and the heterogeneity in users' responsiveness (on subscription) to different trial lengths. 

\subsection{Average Treatment Effect of Trial Lengths}
\label{ssec:average_effect}
In a fully randomized experiment (such as ours), the average effect of a treatment can be estimated by simply comparing the average of the outcome of interest across treatments. We set the 30-day condition as the control and estimate the average effects of the 14 and 7-day treatments on subscriptions for training and test data. The results from this analysis are shown in Table \ref{table:average_comp}. 
\begin{table}[t]
\centering
\footnotesize
\begin{tabular}{ll|cccc}
\toprule
\multirow{2}{*}{\textbf{Data}} & \multirow{2}{*}{\textbf{Treatment}} &   \textbf{Subscription rate} &\textbf{Subscription rate}  &  \multirow{2}{*}{\textbf{t-statistics}} & \textbf{Percentage gain}\\
& & &\textbf{difference}  &  & \textbf{over baseline}\\
\hline\hline
\multirow{3}{*}{Training data}  &    7 days &        0.1532 &                   0.0064 &          3.08 &                4.34 \\
     & 14 days&           0.1490 &                   0.0021 &          1.03 &                1.45 \\
     & 30 days&           0.1468 &                 --- &         --- &               --- \\
    \hline 
\multirow{3}{*}{Test data} & 7 days &           0.1544 &                   0.0082 &            2.58 &                5.59 \\
     & 14 days &           0.1511 &                   0.0048 &          1.51 &                3.28 \\
     & 30 days&           0.1463 &                --- &          --- &                --- \\
     
\bottomrule
\end{tabular}
\caption{\label{table:average_comp} Average effect of the 7- and 14-day treatments on subscription; compared to the control condition of 30-day free-trial. Baseline subscription rate (for 30-day case): $14.68$ in training data and $14.63$ in test data.}
\end{table}

The 7-day trial increases the subscription rate by $4.34\%$ over the baseline of the 30-day condition in the training data and by $5.59\%$ in the test data. However, in both data sets, the effect of the 14-day trial is not significantly different from that of the 30-day trial. These results suggest that a uniform targeting policy that gives the 7-day treatment to all users can significantly increase subscriptions. We also see that the average treatment effect is fairly small compared to the outcome, which is either zero or one. This makes sense because it is natural that the effect of the length of the free trial is small compared to other factors that can affect customer acquisition, especially in a near monopolistic market. Finally, note that the gains and subscription rates in the training and test data are slightly different. As discussed earlier, this is due to the randomness in the splitting procedure.

\subsection{Heterogeneity in User Response}
\label{ssec:het_treatment}

\begin{figure}[!ht]
    \includegraphics[width=160mm]{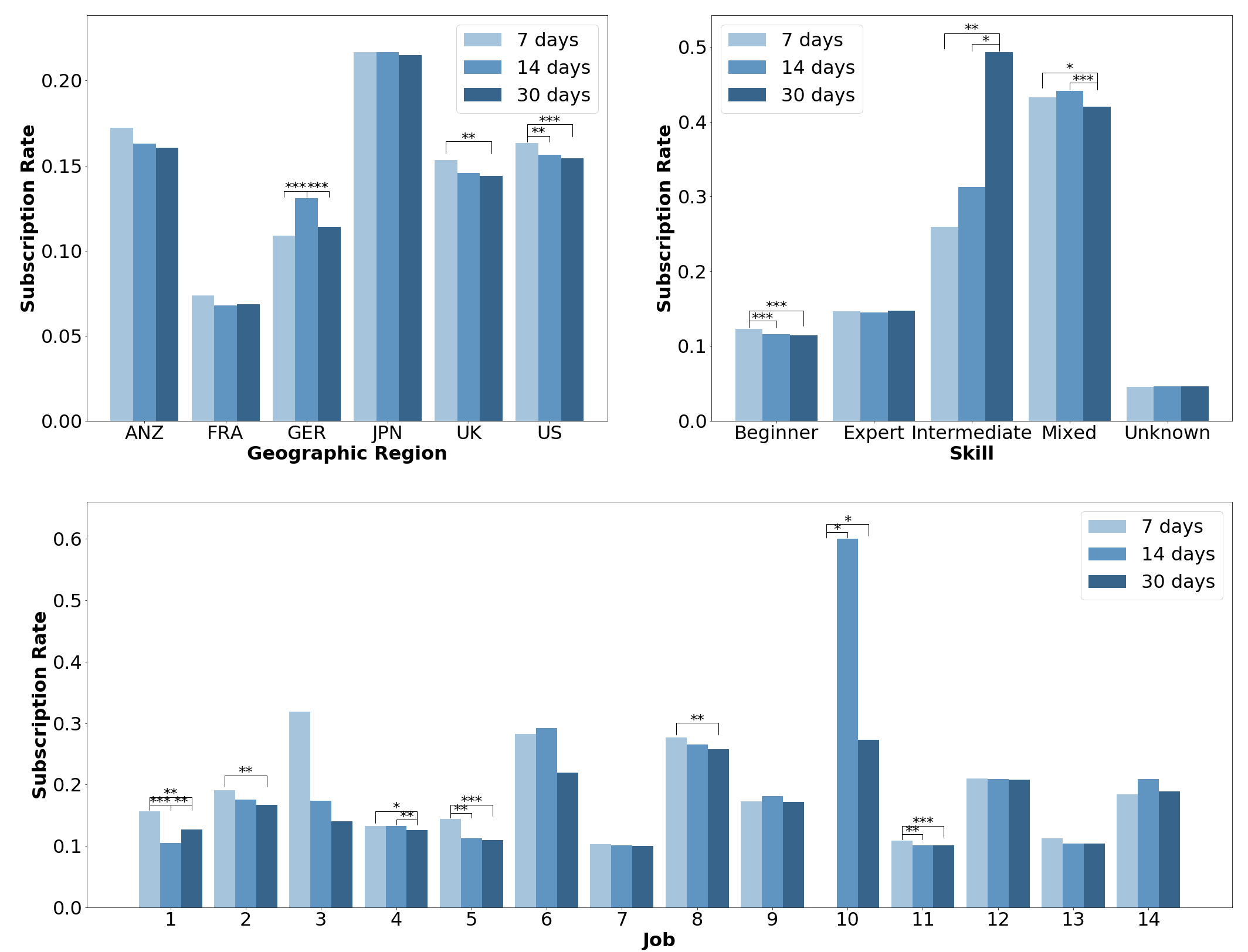}\\
    \centering
    \footnotesize
    * $p < 0.1$, ** $p < 0.05$, *** $p < 0.01$\\
\caption{Heterogeneity in consumers' response to the three trial lengths within three categories -- Geographic region, Skill, and Job. The six geographic regions shown are: Australia and New Zealand, France, Germany, Japan, and United States of America (in that order). We do not include sub-category names for Job to preserve the firm's anonymity.}
\label{figure:HeterBase}
\end{figure}

We now examine whether the firm can do better by personalizing its treatment assignment based on a user's pre-treatment characteristics. In the top left panel of Figure \ref{figure:HeterBase}, we partition the data into six sub-groups based on the user's geographic region and present the average subscription rates for the three trial lengths for each region. The results suggest that there is some heterogeneity in response rates by region. For example, the 14-day trial is more effective in Germany while the 7-day trial is more effective in the United States of America. Next, we perform a similar exercise on skill-level and job (see the top right and bottom panels in Figure \ref{figure:HeterBase}). Again, we find that users' responsiveness to the treatment is a function of their skill level and job. For instance, the 7-day trial is significantly better for Beginners, whereas the 14-day trial is more effective for Mixed-skill users. 

\begin{figure}[t]
    \centering
    \includegraphics[width=120mm]{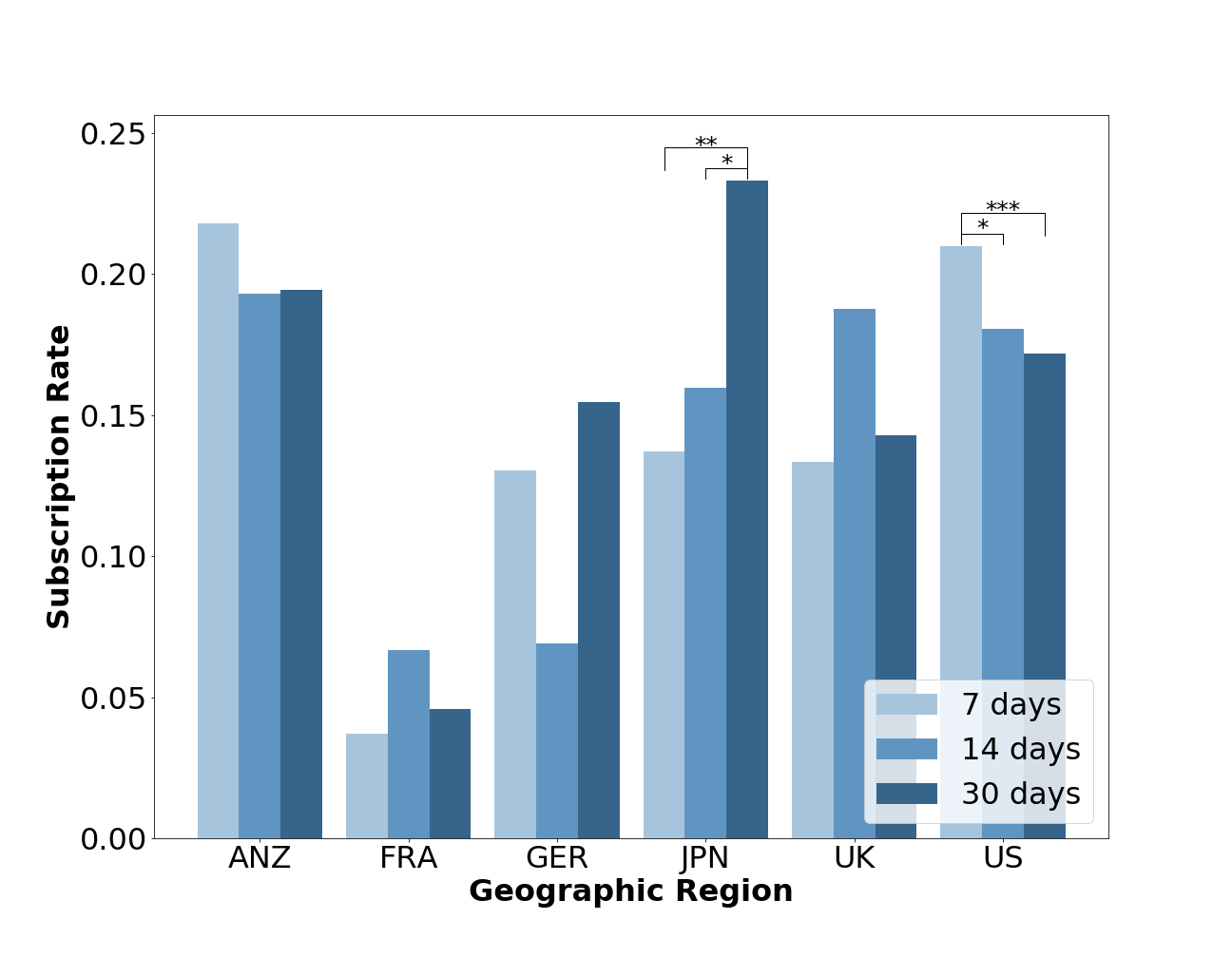}
     \\
    \footnotesize
    * $p < 0.1$, ** $p < 0.05$, *** $p < 0.01$\\
        \caption{\label{figure:interacHeter} Subscription rates for users whose Job $= 2$ across the six geographic regions.}
\end{figure}

We also find that users' responsiveness varies with the interaction of two or more pre-treatment variables. Figure \ref{figure:interacHeter} shows the treatment effect for users whose job description is Business Professionals by their geographic region.  Based on Figure \ref{figure:HeterBase}, we know that, on average, the 7-day trial is better than the 30-day trial for Business Professionals (Job $=2$). However, this effect varies by the user's geographic location. While Business Professionals in the US do better with shorter free trials, those in Japan respond better to the 30-day trial.

Together, these results suggest that users' responsiveness to trial lengths is heterogeneous along many pre-treatment variables and their higher-order interactions. If the firm can successfully exploit this heterogeneity and tailor its free trial assignment policy at the individual-level, then it may be able to achieve higher subscriptions than simply assigning all users to the 7-day trial. 

\section{Personalized Policy Design and Evaluation}
\label{sec:per_pol}
We now present a three-pronged framework that allows us to design and evaluate personalized targeting polices. In $\S$\ref{ssec:opt_pol_des}, we define the optimal personalized policy design problem and solution concept. Next, in $\S$\ref{ssec:estimators}, we discuss the set of estimators that can be used to design the optimal personalized policy. Finally, in $\S$\ref{ssec:pol_eval}, we discuss policy evaluation. Throughout this section, we describe the framework in general terms and discuss the application to our free trials data in $\S$\ref{sec:counter_policies}. 

\subsection{Optimal Policy Design}
\label{ssec:opt_pol_des}
We define the optimal policy design problem in $\S$\ref{sssec:probdef} and discuss the solution concept in $\S$\ref{sssec:sol_concept}.

\subsubsection{Problem Definition}
\label{sssec:probdef}
Consider a setting with $i\in \{1, \ldots, N\}$ independent and identically distributed users, where each user is characterized by a pre-treatment covariate vector $X_i \in X$ of dimension $D$. Let $W_i \in \mathcal{W}$ denote the treatment or intervention that $i$ receives. $\mathcal{W} = \{0, \ldots, W-1\}$ refers to the set of treatments, and the total number of treatments is $W$. We restrict ourselves to settings where treatments are assigned based on a fully randomized experiment or those where the treatment propensities are a function of observable variables (e.g., an observational study). We use $e(W_i = w, X_i)$ to denote the probability that user $i$ with pre-treatment attributes $X_i$ is assigned treatment $w$. Finally, let $Y(X_i, W_i)$ denote the outcome for a user $i$ with pre-treatment variables $X_i$ when she is allocated treatment $W_i$. 

We begin with a formal definition of a personalized treatment assignment policy.
\begin{defn}
\textbf{Personalized treatment assignment policy.} $\pi$ is defined as a mapping between users and treatments such that each user is allocated one treatment, $\pi: X \rightarrow \mathcal{W}$.
\end{defn}
$\pi(X_i) = W_i^{\pi}$ implies that the policy $\pi$ assigns treatment $W_i^{\pi}$ to a user $i$ with pre-treatment attributes $X_i$. 

Next, we define the objective function that a firm or policy-maker wants to maximize. Let $Y$ denote an outcome of interest. Then:
\begin{defn}
\textbf{Reward function.} The firm's  objective is to choose a policy $\pi$ such that it maximizes the expectation of outcomes, $\frac{1}{N}\E \left[\sum_{i=1}^{N}  Y(X_i, W_i^{\pi})\right]$. Thus, for policy $\pi$ and outcome of interest $Y$, we can write our reward function as:
\begin{equation}
R(\pi, Y) = \frac{1}{N} \sum_{i=1}^{N}  \E \left[ Y(X_i, \pi(X_i))\right].
\end{equation}
\end{defn}
We can now define the optimal personalized policy as follows:
\begin{defn}
\textbf{Optimal personalized policy.} Given a reward function $R(\pi, Y)$, the optimal personalized policy is given by: 
\end{defn}
\begin{equation}
    \pi^* =  \argmax\limits_{\pi \in \Pi} \left[R(\pi, Y) \right],
\end{equation}
where $\Pi$ is the set of all possible policies. 

The problem of finding the optimal personalized policy is equivalent to one of finding the policy $\pi^*$ that maximizes the reward function $R(\pi, Y)$. As discussed in $\S$\ref{sec:intro}, this is a non-trivial problem since the cardinality of the policy space can be quite large. The total number of possible policies is $W^{\prod_{d=1}^D c_d}$, when we have $D$ pre-treatment variables and the $d$-th variable can take $c_d$ different values. 

\subsubsection{Solution Concept}
\label{sssec:sol_concept}
Given the cardinality of the policy space, a direct search over the policy space to find the optimal policy is infeasible. So we discuss the two-step approaches available to find the optimal policy $\pi^*$ that avoid this problem. However, we first need to make three assumptions on the data generating process for these approaches to be valid.

\begin{assum}[\emph{Unconfoundedness}]
The treatment assignment is (conditionally) independent of the potential outcome: $W_i \perp \left(Y(W_i=0), \ldots, Y(W_i=W-1) \mid X_i \right )$.
\label{assum_unconf}
\end{assum}
The unconfoundedness assumption holds when we have a fully randomized experiment or when treatment assignment propensities are only function of observables. 

\begin{assum}[\emph{Stable Unit Treatment Value Assumption}]
SUTVA implies that units do not interfere with each other. It requires the response of a user to only depend on the treatment to which she was assigned, and not the treatments of others around her. 
\label{assum_sutva}
\end{assum}
SUTVA is a reasonable assumption in most marketing settings, and it is violated only when there are WOM or network effects. 

\begin{assum}[\emph{Positivity}]
All users have a positive probability of receiving all values of the treatment variable: $e(W_i = w, X_i)> 0$ for all $w \in \mathcal{W}$ and $X_i$.
\label{assum_positivity}
\end{assum} 
This assumption is satisfied in randomized experiments and in observational studies where there is residual randomness in treatment assignment, conditional on observables. This assumption is necessary for both being able to design policy for the full distribution of $X$s over all $W$s, as well as for empirically evaluating the performance of a policy $\pi$ based on the data we have. Intuitively, if there are some combinations of covariates and treatments that have a zero probability of occurring in our data, then we cannot make inferences in this space nor can we evaluate how a counterfactual policy $\pi$ would perform in this space.

With these assumptions in place, we can design the optimal personalized policy if we either have estimates of the outcome of interest or pairwise treatment effects. We discuss both these approaches in detail below.

\squishlist
\item  \textbf{Policy design using outcome estimates:} In this approach, we first obtain consistent estimates of outcomes, $Y(X_i, W_i)$, for all combinations of $w$ and $x$. We can write a statistical model to estimate outcomes as follows:
\begin{equation}
f(x,w)=\E[Y|X_i=x, W_i=w]
\label{eq:out_est}
\end{equation}
As long as Assumptions \ref{assum_unconf} through \ref{assum_positivity} are satisfied, any flexible model $f(x,w)$ will give us consistent estimates of the outcome $y$ for any combination of covariates and treatments. 

Once we have a consistent estimate of the expected outcome, $\hat{y}(x = X_i, w)$, for all possible treatments for a given covariate vector $X_i$, we can obtain the optimal personalized policy as:
\begin{equation}
    \pi^*(X_i) = w^*, \quad \textrm{where} \quad  w^* = \argmax_{w \in \mathcal{W}} \hat{f}(x = X_i, w)
    \label{eq:policy_assign_outcome}
\end{equation}

\item \textbf{Policy design using estimates of pairwise heterogeneous treatment effects:} The second approach to designing a personalized policy is to first obtain consistent estimates of heterogeneous treatment effects for each pair of treatments for all users, and then use them to assign treatments. This method also follows a two-step procedure.

\squishlist

\item In the first step, we can obtain consistent estimates of individual-level treatment effects, $\tau_{w_j, w_{j'}}(x)$, for each pair of treatments $w_j, w_{j'} \in \{0, \ldots, W-1\}$. Under Assumptions \ref{assum_unconf} through \ref{assum_positivity}, $\tau_{w_j, w_{j'}}(x)$, can be defined as \citep{Rubin_1974, Imbens_Rubin_2015}:
\begin{equation}
    \tau_{w_j, w_{j'}}(x) = \E \Big[Y(X_i, W_i = w_j) - Y(X_i, W_i = w_{j'})\mid X_i = x \Big].
    \label{eq:potential_outcome_gen}
\end{equation}
So if we have $W$ treatments, we have to build $\frac{W(W-1)}{2}$ pairwise estimators/models, where each model gives us an estimate of individual-level treatment effects for a given pair of treatments, $w_j, w_{j'}$. 

\item In the second step, we use the estimated treatment effects to derive the optimal policy as:
\begin{equation}
    \pi^*(X_i) = w_j \quad \textrm{if and only if} \quad  \forall \;\; j' \neq j \quad \tau_{w_j, w_{j'}}(x = X_i) \geq 0
    \label{eq:policy_assign_cate}
\end{equation}
In practice, we can have situations where three or more pairs of the estimated treatments form a loop, i.e., $\hat{\tau}_{w_j, w_{j'}}(x = X_i) > \hat{\tau}_{w_j', w_{j''}}(x = X_i) > \hat{\tau}_{w_j'', w_{j}}(x = X_i)$. This happens if the estimated treatment effects are noisy (usually due to lack of sufficient data). For these observations, we can simply assign the best average treatment (across all observations) as the policy-prescribed treatment.\footnote{There can be multiple optimal policies because two or more top treatments can have the same effect on the outcome. That is, for some combinations of $X_i$ and $j' \neq j$, we can have: $\tau_{w_j, w_{j'}}(x = X_i) = 0$. However, notice that all the solutions give the same reward, i.e., the optimal reward is unique.}
\squishend
\squishend

Intuitively, both these approaches try to put some structure on the relationship between a user's responsiveness and her treatment and demographic variables in the first step. Then, in the second step, they use this structure to assign the best policy. This enables them to avoid an unstructured search for the best policy. Both approaches are consistent and correct. With an infinitely large dataset and sufficiently flexible models (of outcomes and treatment effects), both should give the exact same optimal policy. However, in practice, data are finite, the covariate space is high dimensional, and different outcome/treatment effect estimators impose very different parametric/semi-parametric assumptions on users' response behavior. Therefore, the optimal policy in a given empirical setting can vary significantly based on the exact model used to estimate the outcome or heterogeneous treatment effects. Indeed, one of the main objectives of this work is to examine the efficacy of different types of estimators for designing policy.

\subsection{Estimators Used for Policy Design}
\label{ssec:estimators}
We now present the second component of our framework -- the estimators used for policy design. In $\S$\ref{sssec:outcome}, we discuss a series of outcome estimation methods, and in $\S$\ref{sssec:het-effect}, we discuss heterogeneous treatment effects estimators.

As discussed in $\S$\ref{ssec:agenda_challenges}, an empirical challenge that we face when estimating outcomes or heterogeneous treatments effects is the dimensionality of the covariate space of consumers' attributes (which is equal to $\prod_{d=1}^D c_d$, and can be quite high even in simple settings). In most applications, we will therefore not have sufficient data to estimate a local model in each sub-region of the covariate space. We require methods that can automatically pool together observations from sub-regions that show similar responsiveness to the firm's actions. We face a standard bias-variance trade-off here -- models that impose more structure (e.g., regression) may have high in-sample bias if the structure does not match the true data generating process. But they will also have low out-of-sample variance since the model is more global. On the other hand, models that are more flexible (e.g., random forests) may have low in-sample bias (i.e., over-fit to the training data), but perform poorly out-of-sample. Thus, which of these estimators is the best for a given setting is an empirical question. 

\subsubsection{Outcome Estimators}
\label{sssec:outcome}
Outcome estimators seek to learn a model $f(x,w)=\E[Y|X_i=x, W_i=w]$. We consider five commonly used outcome estimators: (1) linear regression, (2) lasso, (3) CART, (4) random forest, and (5) boosted regression trees. We focus on these because of a few key reasons. First, linear regression is the simplest and most commonly used method to model any outcome of interest. Second, lasso is worth exploring because it was designed to improves on the predictions from a linear regression by reducing the out-of-sample variance using variable selection. Third, CART is a semi-parametric way to model outcomes by partitioning the covariate space into areas with the highest within-group similarity in outcomes. Finally, both random forest and boosted regression trees are improvements of CART and have been shown to offer much higher predictive ability than simpler models such as CART, regression, or lasso. We discuss each of these briefly below.

\textbf{Regression-based methods:} A linear regression model with first-order interaction effects can be used to predict an individual $i$'s observed outcome $Y_i$ as a function of her pre-treatment variables, treatment variable, and the interaction of both, as follows:
\begin{equation}
    Y_i =  X_i \beta_1 +  W_i \beta_2 +  X_i W_i \beta_3 + \epsilon_i.
    \label{eq:reg}
\end{equation}
$\beta_1$ is a vector that captures the effect of the pre-treatment variables ($X_i$) on the outcome. $\beta_2$ is a vector that captures the main effect of the different treatments on $Y$. Finally, the vector $\beta_3$ captures the interaction effect of treatment and pre-treatment variables. This interaction term is important because it helps us in personalizing the policy by capturing how the effectiveness of different treatments varies across individuals (as a function of their pre-treatment attributes). 

However, in a high dimensional covariate space, linear regressions with first-order interaction effects will have a large number of explanatory variables. This usually leads to poor out-of-sample fit. That is, such regressions tend to have low bias, but high variance -- they tend to overfit on the training sample but perform poorly in new samples, especially if the data generating process is noisy. Since our goal is to design optimal policies for counterfactual situations, the out-of-sample fit is the only metric of importance.\footnote{We can also add higher-order interaction effects into the regression model. However, we refrain from doing so because it will increase model complexity significantly and exacerbate this problem.}

Lasso addresses the above problem by learning a simpler model that uses fewer variables \citep{Tibshirani_1996, Friedman_etal_2010}. Practically, lasso estimates a linear regression that minimizes the MSE with an additional term to penalize model complexity as shown below:
\begin{align}
     \notag
    (\hat{\beta}_1,\hat{\beta}_2,\hat{\beta}_3) = &\argmin \sum_{i=1}^n \left( Y_i - X_i\beta_1-W_i\beta_2-X_i W_i\beta_3 \right)^2 + \lambda (  ||\beta_{1}||_1+||\beta_{2}||_1+||\beta_{3}||_1 ),
     \label{eq:regression_equation}
\end{align}
where $||\beta_{i}||_1$ is the L1 norm of the vector $\beta_{i}$ and is equal to the sum of the absolute value of the elements of vector $\beta_{i}$. Intuitively, if there are multiple weak (and correlated) predictors, lasso will pick a subset of them and force the coefficients of others to zero, thereby estimating a simpler model. Model-selection in lasso is data-driven, i.e., $\lambda$ is a hyper-parameter that is learned from the data (and not assumed).\footnote{Another regression-based model that is worth considering is elastic net. Elastic net penalizes both L1 norm (like lasso) and L2 norm (like ridge regression). However, in our application setting, the cross-validation procedure chooses the L2 norm coefficient to be zero. So elastic net converges to lasso in our setting. So we do not discuss it in detail here.}

\textbf{Tree-based methods:} Next, we discuss tree-based models, starting with CART. CART recursively partitions the covariate space into sub-regions, and the average of $Y$ in a given region ($E(Y)$) is the predicted outcome for all observations in that region \citep{breiman_etal_1984}. This type of partitioning can be represented by a tree structure, where each leaf of the tree represents an output region. A general CART model can be expressed as:
\begin{equation}
y = f(x,w) = \sum\limits_{m=1}^{M} \rho_m I(x, w \in R_m),
\label{eq:cart}
\end{equation}
where $x, w$ is the set of explanatory variables, $R_m$ is the $m^{th}$ region of the $M$ regions used to partition the space, $\rho_m$ is the predicted value of $y$ in region $m$. 

Trees are trained by specifying a cost function (e.g., mean squared error) that is minimized at each step of the tree-growing process using a greedy algorithm. To avoid over-fitting, usually a penalty term is added to the cost function, whose weight is a hyper-parameter learnt from the data using cross-validation.  

While CART has some good properties\footnote{CART can accept both continuous and discrete explanatory variables, is not sensitive to the scale of the variables, and allows any number of interactions between features \citep{murphy_2012}. It therefore has ability to capture rich nonlinear patterns in the data. Further, CART can do automatic variable selection, i.e., it uses only those variables that provide better accuracy in the prediction task for splitting.}, it often has poor predictive accuracy because it is discontinuous in nature and is sensitive to outliers. Therefore, even with regularization, CART models tend to over-fit on the training data and under-perform on out-of-sample data. We can address these problems using two different techniques -- (1) Bagging and (2) Boosting. We discuss both of these techniques and the resulting estimators below.

In general, deep trees have high in-sample fit (low bias), but high out-of-sample variance because of over-fitting. However, we can improve the variance problem by bagging or averaging deep trees using bootstrapping. \citet{Ho_1995} formalized this idea and proposed random forests. Random forest usually consists of hundreds or thousands of trees, each of which is trained on a random sub-sample of columns and rows. Each tree is thus different from other trees and the average of these random trees is a better predictor than one single tree trained on the full data. 

Another approach is to start with shallow trees. Shallow trees have poor in-sample fit (high bias), but low out-of-sample variance. Additively, adding a series of weak trees that minimize the residual error at each step by a process known as boosting can improve the bias problem, while retaining the low variance. This gives us boosted regression trees. Conceptually, boosting can be thought of as performing gradient descent in function space using shallow trees as the underlying weak learners \citep{Breiman_1998, Friedman_2001}.\footnote{It should be noted that bagging is not a modeling technique; it is simply a variance reduction technique. Boosting, however, is a method to infer the underlying model $y = f(x,w)$. Thus, they are conceptually completely different.} In this paper, we use XGBoost, a version of boosted trees proposed by \citet{Chen_Guestrin_2016} because it is superior to earlier implementations both in terms of accuracy and scalability. 

Please see Appendix $\S$\ref{sec:app_hyperopt} for the set of hyper-parameters that need to be tuned in CART, random forest, and XGBoost.

\subsubsection{Heterogeneous Treatment Effect Estimators}
\label{sssec:het-effect}
These methods are based on the potential outcomes framework and estimate the treatment effect for any pair of treatments ($w_j, w_{j'}$) at each point $x$ as shown in Equation \eqref{eq:potential_outcome_gen}. However, this equation is not very useful in practical settings (where the covariate space is high-dimensional and data are finite) because we would not have sufficient observations at each $X_i$ to estimate precise treatment effects. Therefore, the general idea behind modern heterogeneous treatment effects estimators is to pool observations that are close in the covariate space and estimate the conditional average treatment effect for sub-populations instead of estimating the treatment effect at each point in the covariate space. That is, we can modify Equation \eqref{eq:potential_outcome_gen} as:
\begin{equation}
    \tau_{w_j, w_{j'}}(x) = \frac{\sum_{X_i \in l(x), W_i = w_j}Y_i}{\sum 1[X_i \in l(x), W_i = w_j]} - \frac{\sum_{X_i \in l(x), W_i = w_{j'}}Y_i}{\sum 1[X_i \in l(x), W_i = w_{j'}]},
    \label{eq:potential_outcome_lx}
\end{equation}
where $l(x)$ is the set of covariates that are fairly similar to $x$. Intuitively, for each point $x$, we use the observations in $l(x)$ to estimate treatment effects.

The main question in these methods then becomes how to find the optimal $l(x)$ around each $x$. On the one hand, if $l(x)$ is too small, then we will not have sufficient observations within $l(x)$, which would result in noisy estimates of treatment effects. On the other hand, if $l(x)$ is too large, then we will not capture all the heterogeneity in the treatment effects, which is essential for personalizing policy. Indeed, if $l(x)$ is the entire data, then we simply have one Average Treatment Effect for all users (which will give us one global policy). Thus, finding the optimal $l(x)$ involves the classic bias-variance trade-off. 

Starting with \citet{Rzepakowski_Jaroszewicz_2012}, a growing stream of literature has focused on developing data-driven approaches to finding the optimal $l(x)$ based on ideas from the machine learning literature. Among these methods, the recently developed causal tree and causal forest (or Generalized Random Forest) have been shown to have superior performance. So we focus on them.

Causal tree builds on CART. The only difference is that instead of partitioning the space to maximize predictive ability, the objective here is to identify partitions with similar within-partition treatment effect. \citet{Athey_Imbens_2016} show that maximizing the variation in the estimated treatment effects (with a regularization term added to control complexity) achieves this objective. Their algorithm consists of two steps. In the first step, it recursively splits the covariate space into partitions. In the second step, it estimates the treatment effect within each partition ($l(x)$) using Equation \eqref{eq:potential_outcome_lx}. Intuitively, this algorithm pools observations with similar treatment effects into the same partition because splitting observations that have similar treatment effects does not increase the objective function (i.e., variation in the post-split treatment effect). 

The causal tree algorithm nevertheless suffers from weaknesses that mirror those of CART. To resolve these issues, \citet{Wager_Athey_2018} proposes causal forest, which builds on random forests \citep{Breiman_2001}. More broadly, \citet{Athey_etal_2019} show that the intuition from random forests can be used to flexibly estimate any heterogeneous quantity, $\theta(x)$, from the data, including heterogeneous treatment effects. They suggest a Generalized Random Forest (GRF) algorithm that learns problem-specific kernels for estimating any quantity of interest at a given point in the covariate space. For treatment estimation, the method proceeds in two steps. In the first step, it builds trees whose objective is to increase the variation in the estimated treatment effects. Each tree is built on a random sub-sample of the data and random sub-sample of covariates. In the second step, the algorithm uses the idea of a weighted kernel regression to calculate the treatment effect at each point $x$ using weights from the first step. 

Please see Appendix \ref{sec:app_het_treat} for a detailed discussion both causal tree and causal forest and Appendix $\S$\ref{sec:app_hyperopt} for details on hyper-parameter optimization in these two methods.\footnote{\citet{Athey_Imbens_2016} propose an additional approach to avoid over-fitting in causal tree and causal forest -- honest splitting.  The main idea behind honesty is to split the data into two parts and use one part for growing each tree (i.e., generating the partitions) and the other part for estimating the treatment effects given a partition. Since the two data-sets are independent of one another, there is a lower likelihood of over-fitting. However, honest splitting comes with its own costs -- it reduces the amount of data we have for learning in both stages by half. In settings where the magnitude of the treatment effects is small (such as ours), honest splitting can  adversely affect the performance of models. Indeed, we found that models based on honest splitting lead to worse policies compared to models without honest splitting in our setting. So we do not employ honesty in our analysis.}

\subsection{Policy Evaluation}
\label{ssec:pol_eval}
We now discuss the final component of our three-pronged framework -- offline policy evaluation. 

To identify the optimal policy among the set of all possible policies ($\Pi$), it is important to be able to consistently evaluate the gain from implementing any policy $\pi$ {\it without} actually deploying it. As discussed in $\S$\ref{sec:intro}, this is important because deploying a policy in the field is costly in time and money. Moreover, when the policy space is large, it is simply not feasible to test each policy in the field. So we turn to off-policy estimators proposed in the reinforcement learning literature \citep{Sutton_Barto_2018}, where the goal is to estimate the performance of a given policy $\pi$ using data generated by an experiment or another policy.

\begin{figure}[t]
    \centering
    \includegraphics[scale = 0.37]{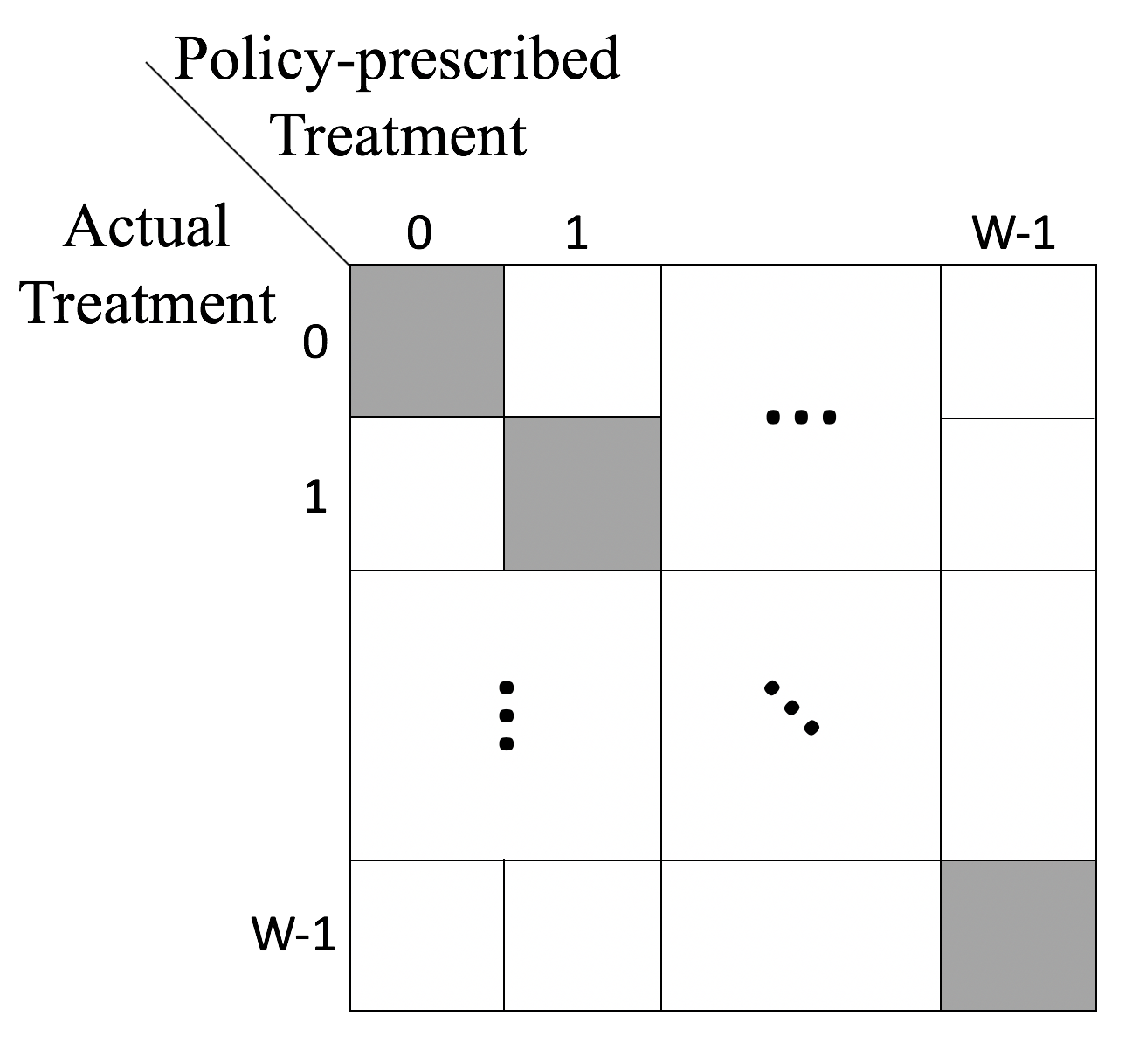}\\
        \caption{\label{figure:actual_policy} Classification of observations based on the actual treatment they received in the data and the treatment prescribed by the policy. The shaded cells in the diagonal constitute policy congruent observations and the white cells constitute policy incongruent observations.}
\end{figure}

In any setting where Assumption \ref{assum_positivity} is satisfied, there is a non-zero probability that a user $i$ will be assigned to the treatment prescribed by a given policy $\pi$. Thus, as shown in Figure \ref{figure:actual_policy}, we have two types of observations in the data:
\squishlist
\item \textbf{Policy congruent observations:} Those for whom the observed treatment assignment ($W_i$) matches the  policy-prescribed treatment ($\pi(X_i)$), i.e., $\pi(X_i) = W_i$. 
\item \textbf{Policy incongruent observations:}  Those for whom the observed treatment assignment is different from the policy-prescribed treatment, i.e., $\pi(X_i) \neq W_i$
\squishend
These two sets of observations form the basis of our evaluation metric: the Inverse Propensity Score (IPS) Estimator for Reward. This estimator has been extensively used in the off-policy evaluation  literature \citep{Horvitz_Thompson_1952, Dudik_etal_2011}, and can be defined as: 
\begin{equation}
 \hat{R}_{IPS}(\pi, Y) = \frac{1}{N} \sum_{i=1}^N \frac{1[{W_i = \pi(X_i)}] Y_i}{e(W_i, X_i)}.
     \label{eq:ips_reward}
\end{equation}
The IPS estimator is unbiased as long as Assumption \ref{assum_positivity} is satisfied. The intuition behind this estimator can be easily understood using Figure \ref{figure:actual_policy}. Essentially, it takes all the observations in the policy congruent cells and scales them up by their propensity of receiving the treatment assigned to them. This scaling gives us a pseudo-population that received the policy-prescribed treatment. Thus, the average of the outcome for this pseudo-population gives us an estimate of the reward for the true population, if we were to implement the proposed policy in the field.

\section{Counterfactual Free Trial Policies}
\label{sec:counter_policies}
We now apply the framework presented above to our free trial experiment and discuss the main findings. We first train the outcome and heterogeneous treatment effects models and discuss their performance in $\S$\ref{ssec:first-stage}. Next, in $\S$\ref{ssec:policies}, we define and derive a set of counterfactual targeting policies, and present the fraction of users assigned the three treatments under each of these policies. Then, we quantify the gains in subscription from adopting different counterfactual policies in $\S$\ref{ssec:gains}, and we finally discuss the source of these differences in $\S$\ref{ssec:source}. 

\subsection{First-stage Estimation and Results}
\label{ssec:first-stage}

We start by briefly describing our implementation of the estimators discussed in $\S$\ref{ssec:estimators}. First, in our setting, all the pre-treatment variables, $X_i$s, are categorical. So we transform each categorical variable into a set of dummy variables for each sub-category (also referred to as one-hot encoding in the machine learning literature). After this transformation, we have 82 dummies for the pre-treatment variables, three treatment dummies, and 246 first-order interaction variables. This gives us a total of 331 explanatory variables for the linear regression and lasso models. For the rest of the tree-based models, we directly feed in the 82 dummy variables for the pre-treatment characteristics and the three treatment dummies.\footnote{The GRF algorithm estimates outcomes and propensities in the first step (see Appendix \ref{ssec:app_grf}). However, in our setting, we do not need to estimate propensity scores since they are known and constant for each observation since treatment assignment is fully randomized. Our qualitative results remain the same if we instead estimate them from the data.} Next, we use five-fold cross-validation on the training data to tune hyper-parameters for all the estimators (except regression which does not require hyper-parameter tuning). See Appendix $\S$\ref{sec:app_hyperopt} for more details on model selection.

The main result of interest for outcome prediction models is their predictive performance. So in Table $\S$\ref{table:predictive-perf}, we present the MSE for the five outcome estimators on both training and test data. Two main findings emerge. First, we find that linear and lasso are the worst in terms of model-fit, with XGBoost performing the best. This finding is consistent with earlier papers that have found XGBoost to be the best outcome prediction method in tasks involving prediction of human behavior \citep{Rafieian_Yoganarasimhan_2020, Rafieian_2019a}. However, we also find that all the tree-based models -- CART, Random Forest, and XGBoost -- suffer from overfitting in spite of hyper-parameter tuning. That is, their performance on the test data is worse than that on training data. This is common in most machine learning models and stems from the differences in the test and training data.

\begin{table}[t]
\centering
\footnotesize

\begin{tabular}{l|cc}
\toprule
\multirow{2}{*}{\textbf{Method}} & 
\multicolumn{2}{c}{\textbf{Mean Squared Error}}\\
 & Training Set & Test Set
\\
\hline\hline

 Linear Regression &    0.0932 &    0.0933 \\
             Lasso &    0.0933 &    0.0933 \\
              CART &    0.0916 &    0.0920 \\
     Random Forest &    0.0904 &    0.0915 \\
           XGBoost &    0.0905 &    0.0911 \\

\bottomrule
\end{tabular}
\caption{\label{table:predictive-perf} Comparison of the predictive performance of the five outcome estimation methods. The MSE for any method are calculated as $\frac{\sum_{i = 1}^N(\hat{y}_i-y_i)^2}{N}$, where $\hat{y}_i$ is the prediction of $y_i$ and $N$ is the number of data-points in the data-set being considered.}
\end{table}

Next, we discuss the results from the heterogeneous treatment effects estimators. In this case, we cannot compare the estimates of treatment effects with any ground truth since we (as researchers and managers) do not know the true treatment effects. So we simply present the distributions of treatment-effects in Appendix \ref{sec:app_cate_est}. 

\subsection{Uniform and Personalized Policies}
\label{ssec:policies}
We start by designing the following three uniform (one length for all) policies:
\squishlist
\item $\pi_{30}$ -- This policy prescribes the 30-day treatment for all users. It was used by the firm at the time of the experiment and we therefore use it as the baseline policy in all our comparisons.
\item $\pi_{14}$ -- This policy prescribes the 14-day treatment for all users.
\item $\pi_{7}$ -- This policy prescribes the 7-day treatment for all users. Since we found that 7 days is the best average treatment in $\S$\ref{ssec:average_effect}, this should be the best uniform policy.
\squishend

Next, we use the framework described in $\S$\ref{sec:per_pol} to design a series of personalized policies. However, before doing so, we need to ensure that Assumptions \ref{assum_unconf} -- \ref{assum_positivity} are satisfied. So we discuss their applicability to our context below:
\squishlist
\item Assumption \ref{assum_unconf} is automatically satisfied because we have a fully randomized experiment.
\item Assumption \ref{assum_sutva} is satisfied because we do not expect any network effects in our setting (since the experiment was run on unconnected users distributed all over the world). 
\item Assumption \ref{assum_positivity} is satisfied in all randomized experiments by definition.
\squishend

We design seven personalized policies based on the estimators discussed in $\S$\ref{ssec:estimators}: $\pi_{reg}$, $\pi_{lasso}$, $\pi_{cart}$, $\pi_{r\_forest}$, $\pi_{xgboost}$,  $\pi_{c\_tree}$, and $\pi_{c\_forest}$. To design these policies, we follow the two optimal policy design methods discussed in $\S$\ref{sssec:sol_concept}. As discussed there, in reality, there is only one optimal policy. In theory, with sufficiently large data and a fully flexible model, we should be able to correctly identify this reward-maximizing optimal policy. However, in practice, each of these methods imposes different semi-parametric restrictions on the functional form of the model and make different bias-variance trade-offs. Thus, with finite data, these policies can vary substantially.

\begin{table}[t]
\centering
\footnotesize
\begin{tabular}{c|cccccccccc}
\toprule
\multirow{1}{*}{\textbf{Policy-prescribed}}& 
\multirow{2}{*}{$\pi_{7}$}&
\multirow{2}{*}{$\pi_{14}$}&
\multirow{2}{*}{$\pi_{30}$}&
\multirow{2}{*}{$\pi_{reg}$}&
\multirow{2}{*}{$\pi_{lasso}$}&
\multirow{2}{*}{$\pi_{cart}$}&
\multirow{2}{*}{$\pi_{r\_forest}$}&
\multirow{2}{*}{$\pi_{xgboost}$}&
\multirow{2}{*}{$\pi_{c\_tree}$}&
\multirow{2}{*}{$\pi_{c\_forest}$}\\
\multirow{1}{*}{\textbf{Treatment}}&&&&&&&&&&\\\hline

\hline\hline
7 Days & 1 & 0 & 0 &  {0.521}& {0.689} & 1  &    {0.444}  & {0.697}   &   1  &  {0.911}  \\

14 Days & 0 & 1 & 0 &  {0.322}&   {0.232} &  0& {0.269} &  {0.185} & 0 &  {0.089}\\

30 Days & 0 & 0 & 1&  {0.157} &   {0.079} & 0&   {0.287} &   {0.118} &   0 &   
   {0}  \\
\hline
\textbf{Total} & 1 & 1 & 1 & 1 & 1 & 1 & 1 & 1 & 1 & 1\\
\bottomrule
\end{tabular}
\caption{\label{table:method_length_ratio} Fraction of users assigned the 7-, 14-, and 30-day trials under counterfactual policies (in test data).}
\end{table}

Table \ref{table:method_length_ratio} presents the fraction of users who are given each of the three treatments (7, 14, and 30 days) under the policies described above. There are a few points worthy of note here. First, policies based on CART and causal tree do not personalize treatment assignment and end up giving the 7-day treatment to all users. Thus, they are equivalent to the best uniform policy: $\pi_{7} \equiv \pi_{cart} \equiv \pi_{c\_tree}$. Second, the two policies based on the two outcome estimators -- lasso and XGBoost -- are somewhat similar in their treatment assignment. Both prescribe the 7-day treatment to $\approx 70\%$ of users, the 14-day treatment to $\approx 20\%$ of users, and the 30-day treatment to $\approx 10\%$ of users. In contrast, $\pi_{reg}$ and $\pi_{r\_forest}$ prescribe the 7-day treatment to the least number of users while $\pi_{c\_forest}$ prescribes the 7-day treatment to 91\% of users (and the 30-day treatment to no one). Thus, a main takeaway here is that each of these policies behaves quite differently when it comes to treatment assignment.

\subsection{Gains from Counterfactual Policies}
\label{ssec:gains}

\subsubsection{Empirical Policy Evaluation}
\label{sssec:emp_pol_eval}
We can evaluate the expected reward from each policy described using Equation \eqref{eq:ips_reward}. In theory, in a randomized experiment, the propensity of treatment assignment is orthogonal to the treatment prescribed by any policy $\pi$. Thus, $e(W_i = w, X_i) = e(W_i = w) \; \forall \; w \; \in \mathcal{W}$ is known and constant for all observations. However, in practice, within the set of users for whom policy $\pi$ prescribes $w$, the empirical treatment propensities ($\hat{e}_{\pi(X_i)}(W_i=w)$) might not be the same as that in the full data. Thus, to correctly estimate the reward under a given policy, we need to use the empirical treatment propensities within each assigned trial length. We therefore modify Equation \eqref{eq:ips_reward} as follows:
\begin{equation}
 \hat{R}_{IPS}(\pi, Y) = \frac{1}{N} \sum_{i=1}^N \frac{1[{W_i = \pi(X_i)}] Y_i}{\hat{e}_{\pi(X_i)}(W_i)},
     \label{eq:ips_reward_emp}
\end{equation}
where ${\hat{e}_{\pi(X_i)}(W_i)}$ is the probability that a user whom the policy prescribes treatment $\pi(X_i)$ is given $W_i$. Formally,  ${\hat{e}_{\pi(X_i)}(W_i)}= \frac{\frac{1}{N}\sum_{j=1}^N 1[W_j=W_i,\pi(X_j)=\pi(X_i)]}{\frac{1}{N}\sum_{j=1}^N 1[\pi(X_j)=\pi(X_i)]}$. Thus, instead of calculating the treatment propensities on the entire data, we now do it on the subset of users who are prescribed $\pi(X_i)$ by the policy $\pi$. Intuitively, if we look at Figure \ref{figure:actual_policy}, instead of calculating treatment propensities on the full square (i.e., all the data), we calculate them separately for each column (i.e., set of users for whom the policy prescribes a treatment). 

We present the expected reward from all the policies in Table \ref{table:method_comparison_sub}. In all our reward calculations, we use subscription as the outcome of interest. Thus, the expected rewards shown in columns 3 and 4 are the estimated subscription rates (in \%) for the training and test data, respectively.\footnote{Estimates of the subscription rates (or reward) with theoretical propensities based on Equation \eqref{eq:ips_reward} are very similar to those shown in Table \ref{table:method_comparison_sub}.}
 
We clarify a few points before discussing the results. First, all the results will look more promising on the training data since the policies developed are based on models trained and validated on this data. So, the test data provide a more neutral ground for model comparisons and we focus on it in our discussions. That said, we cannot directly compare the estimated subscription rates (or expected reward) for different datasets \citep{Yi_etal_2013}. In general, the performance of a given policy or model on a dataset is a function of: (a) the model used to form the policy, and (b) the joint distribution of the covariates and outcomes in that dataset. Simply put, in Table \ref{table:method_comparison_sub}, comparisons within a column are valid but comparisons across columns are less meaningful. 

\begin{table}[t]
\centering
\footnotesize

\begin{tabular}{ll|cc|cc}
\toprule
\multirow{2}{*}{\textbf{Policy category}} & \multirow{2}{*}{\textbf{Policy}} & 
\multicolumn{2}{c}{\textbf{Estimated Subscription (\%)}}&
\multicolumn{2}{c}{\textbf{Increase in subscription (\%)}}\\

& & Training Set & Test Set & Training Set & Test Set
\\
\hline\hline
\multirow{3}{*}{\shortstack[l]{One length  \\ for all}} & $\pi_{30}$ (Baseline) & 14.68 & 14.63 & --- & ---  \\
& $\pi_{14}$ & 14.90 & 15.11 & 1.45 & 3.28\\
& $\pi_{7}$  & 15.32 & 15.44 & 4.34 & 5.59\\
\midrule
\multirow{5}{*}{\shortstack[l]{Based on \\outcome \\ estimation}}& $\pi_{reg}$  & 15.89 & 15.33 & 8.21 & 4.83\\
& $\pi_{lasso}$  & 15.85 & 15.62 & 7.97 & 6.81\\
& $\pi_{cart}$ & 15.32 & 15.44 & 4.34 & 5.59\\
& $\pi_{r\_forest}$  & 17.42 & 14.82 & 18.67 & 1.32\\
& $\pi_{xgboost}$  & 16.00 & 15.53 & 8.98 & 6.17\\
\midrule
\multirow{2}{*}{\shortstack[l]{Based on treatment \\ effect estimation}} & $\pi_{c\_forest}$  & 15.58 & 15.46 & 6.09 & 5.71\\
& $\pi_{c\_tree}$  & 15.32 & 15.44 & 4.34 & 5.59\\
\bottomrule
\end{tabular}
\caption{\small\label{table:method_comparison_sub} Gains in subscription from implementing different counterfactual free-trial policies. The results for policies $\pi_{cart}$, $\pi_{c\_tree}$, and $\pi_{7}$ are the same since they prescribe the 7-days treatment to all users.}
\end{table}

\subsubsection{Comparison of Counterfactual Policies}
\label{sssec:comp_counter}

The top four policies based on Table \ref{table:method_comparison_sub} are: $\pi_{lasso}$, $\pi_{xgboost}$, $\pi_{c\_forest}$, and $\pi_{7}$, in that order. However, these differences may not be significant. Therefore, in Table \ref{table:paired_t_test_sub}, we present results from paired t-tests comparing the IPS-reward estimates for the top four methods with each other and show that all these differences are significant. We refer readers to Appendix $\S$\ref{sec:app_ttest} for details on these tests.

We now discuss the main findings of interest from Tables \ref{table:method_comparison_sub} and \ref{table:paired_t_test_sub}. First, we find that 7-days for all is the best uniform policy (5.59\% increase over the baseline of 30-days for all).

Next, we find that among the personalized policies, the policy based on lasso is the most profitable (6.81\% improvement in subscription rate), followed by the policy based on XGBoost (6.17\% improvement in subscription rate). Moreover, the differences between these policies and others are significant, i.e., the reward from $\pi_{lasso}$ is significantly better than that from other policies (see column 2 in Table \ref{table:paired_t_test_sub}). Similarly, $\pi_{xgboost}$ is significantly better than all other policies, except $\pi_{lasso}$ (see column 3 in Table \ref{table:paired_t_test_sub}).

However, not all outcome estimation methods do well when it comes to policy design. Linear regression and random forest overfit to the training data (high estimated subscription in-sample), but perform poorly on test data.\footnote{While this could be due to differences in the data itself, the magnitude of this difference suggests overfitting.} Interestingly, we do not find much correlation between an outcome estimator's predictive ability and its efficacy in policy design (recall Table \ref{table:predictive-perf}). This is expected since the objective function in outcome estimation methods is predictive ability, which is different from policy design or performance.

An important point of note here is that poorly designed personalized policies (e.g., those based on regression and random forest) can actually do worse than the best uniform policy based on average treatment effects. As shown in the last column of Table \ref{table:paired_t_test_sub}, the reward from $\pi_7$ is significantly better than that from the personalized policies, $\pi_{reg}$ and $\pi_{r\_forest}$. This problem becomes obvious only when we look at the evaluation metric (IPS reward estimate) on the test data. Simply looking at the relative performance of these policies in the training data can lead to the (incorrect) conclusion that $\pi_{reg}$ and $\pi_{r\_forest}$ outperform the uniform policy $\pi_7$. We, as policy-makers, should therefore be careful in both designing and evaluating personalized policies. It is essential that: (1) we do not conflate a model's predictive ability with its ability to form policy, and (2) we evaluate the performance of each policy on an independent test data with appropriate policy evaluation metrics.

\begin{table}[t]
\centering
\footnotesize

\begin{tabular}{l|cccc} 
\toprule
Compared & 
\multirow{2}{*}{$\Delta \hat{R} (\pi_{lasso}, \pi)$} &  
\multirow{2}{*}{$\Delta \hat{R} (\pi_{xgboost}, \pi)$} & 
\multirow{2}{*}{$\Delta \hat{R} (\pi_{c\_forest}, \pi)$}&
\multirow{2}{*}{$\Delta \hat{R} (\pi_{7}, \pi)$} \\ 
Policy ($\pi$) &  & & & \\

\hline\hline
$\pi_{xgboost}$ & 0.88 (2.32 $\times$ $10^{-28}$) & --- & --- &  --- \\
$\pi_{c\_forest}$ & 1.49 (5.24 $\times$ $10^{-65}$) & 0.62 (8.88 $\times$ $10^{-18}$) & --- & --- \\
$\pi_7$ & 1.72 (3.36 $\times$ $10^{-72}$) &         0.84 (1.73 $\times$ $10^{-27}$) & 0.23 (6.02 $\times$ $10^{-08}$) & --- \\
$\pi_{reg}$ & 2.82 ($\leq$ $10^{-100}$) &               1.94 ($\leq$ $10^{-100}$) & 1.32 (6.36 $\times$ $10^{-54}$) & 1.1 (4.43 $\times$ $10^{-33}$) \\
$\pi_{r\_forest}$ & 7.94 ($\leq$ $10^{-100}$) & 7.06 ($\leq$ $10^{-100}$) & 6.45 ($\leq$ $10^{-100}$) & 6.22 ($\leq$ $10^{-100}$) \\
\bottomrule
\end{tabular}
\caption{\small\label{table:paired_t_test_sub} Comparison of reward estimates based subscription for top four policies with each other and other counterfactual policies (on test data). $\Delta \hat{R} (\pi_{a}, \pi_{b})$ is defined as  $\hat{R}_{IPS}(\pi_{a}, Y) - \hat{R}_{IPS}(\pi_{b}, Y)$. For each cell, we show $\Delta \hat{R} (\pi_{a}, \pi_{b})$ multiplied by $10^3$ and the $p$-value for the paired t-test in parenthesis.}
\end{table}

Finally, we find that the recently proposed heterogeneous treatment effects estimators, causal tree and causal forest, perform poorly when it comes to personalized policy design. As discussed in $\S$\ref{ssec:policies}, causal tree does not personalize the policy at all since it assigns the 7-day trial to all users. While $\pi_{c\_forest}$ does personalize treatment assignment to some degree, the gain over the simple 7-days for all policy is very marginal. (As shown in column 4 of Table \ref{table:paired_t_test_sub}, $\Delta \hat{R} (\pi_{c\_forest}, \pi_7)$ is positive and significant, but very small compared to the gain from other methods such as lasso and XGBoost.) Thus we have: $\hat{R}_{IPS}(\pi_{c\_tree}, Y) = \hat{R}_{IPS}(\pi_{7}, Y) \approx \hat{R}_{IPS}(\pi_{c\_forest}, Y)$. 

Our results thus suggest that managers may be better off adopting the best uniform policy instead of investing resources in personalizing policies based on these methods. This is a useful finding since these methods are gaining traction in the marketing literature and researchers are starting to use them (e.g., \citet{Guo_etal_2017}, \citet{Fong_etal_2019}). However, both in marketing research and practice, the end goal is often policy design and not estimation of treatment effects (e.g., decide who should get a price promotion and who should not, or who should be targeted with advertising and who should not be). In such cases, relying on heterogeneous treatment effects estimators, such as causal forest, can be sub-optimal. Further, the causal forest model suffers from a lack of interpretability and is more susceptible to hyper-parameter tuning issues. In contrast, lasso is not only easy to specify, estimate, and interpret but also offers the best performance for policy design (at least in our setting).

\subsubsection{Policy Performance and CATE Estimates}
\label{sssec:pol_per_cate}
\begin{figure}[!htbp]
    \centering

    \includegraphics[scale = 0.26]{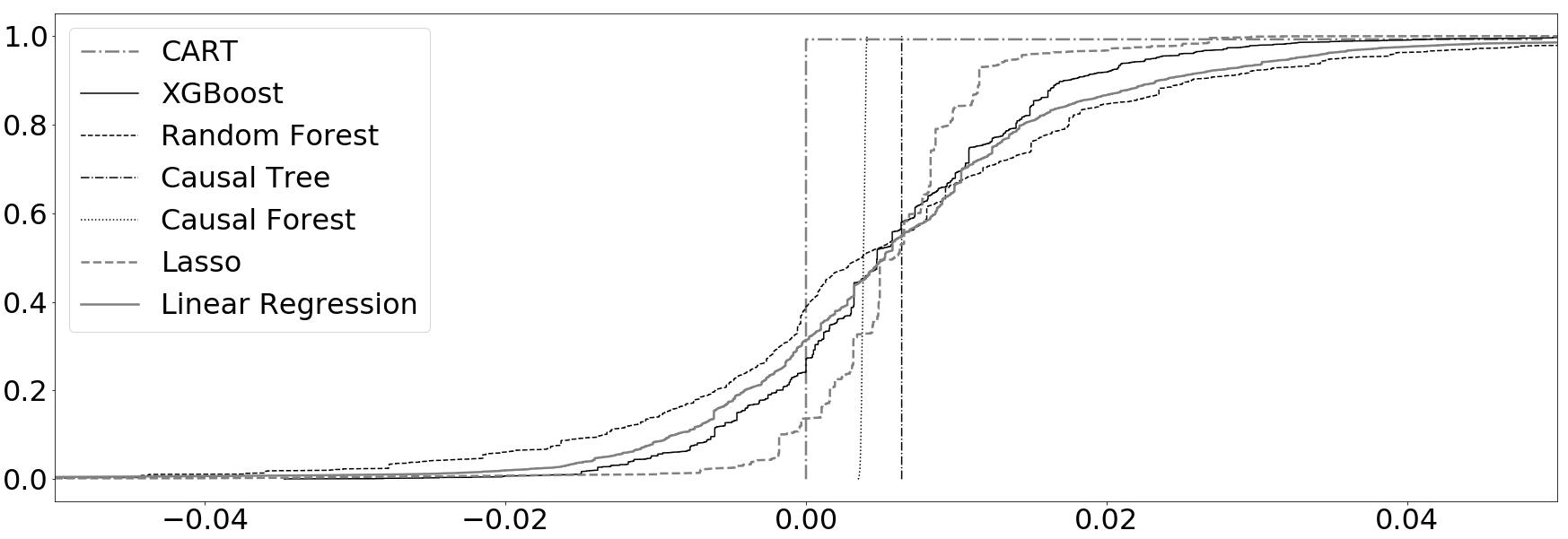}
    CDF of $\mathbf{\tau_{7, 30}}$\par\medskip
    
    \includegraphics[scale = 0.26]{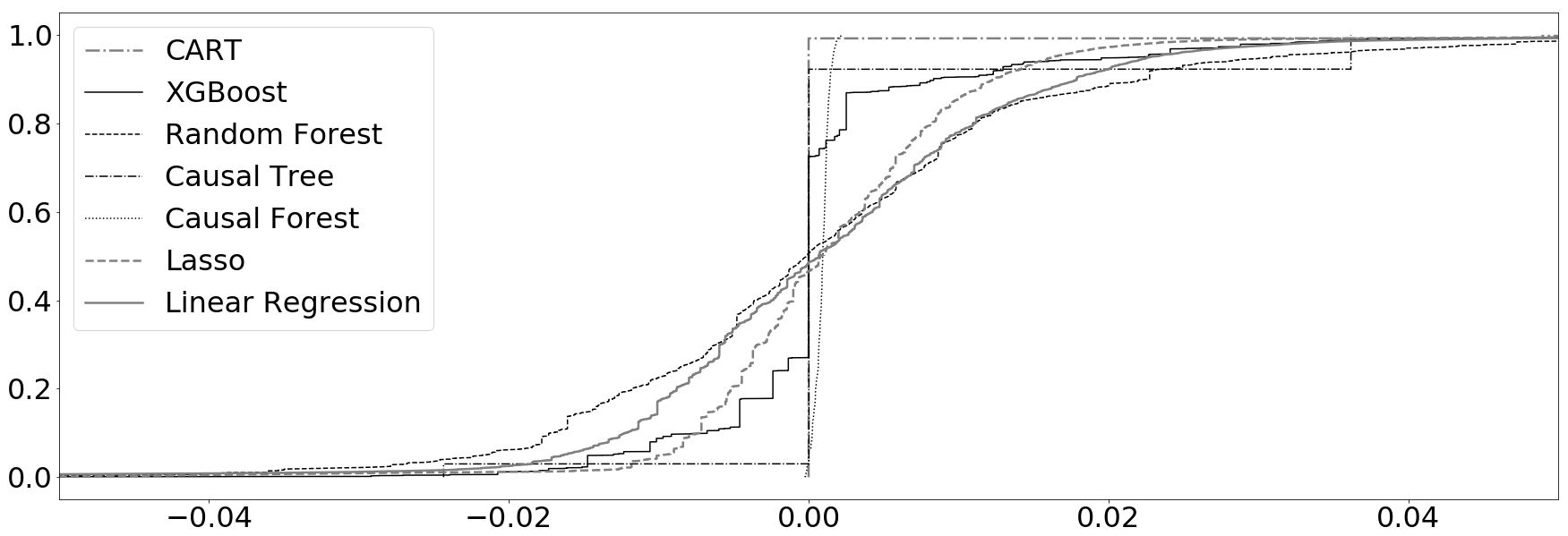}
    CDF of $\mathbf{\tau_{14, 30}}$\par\medskip
    
    \includegraphics[scale = 0.26]{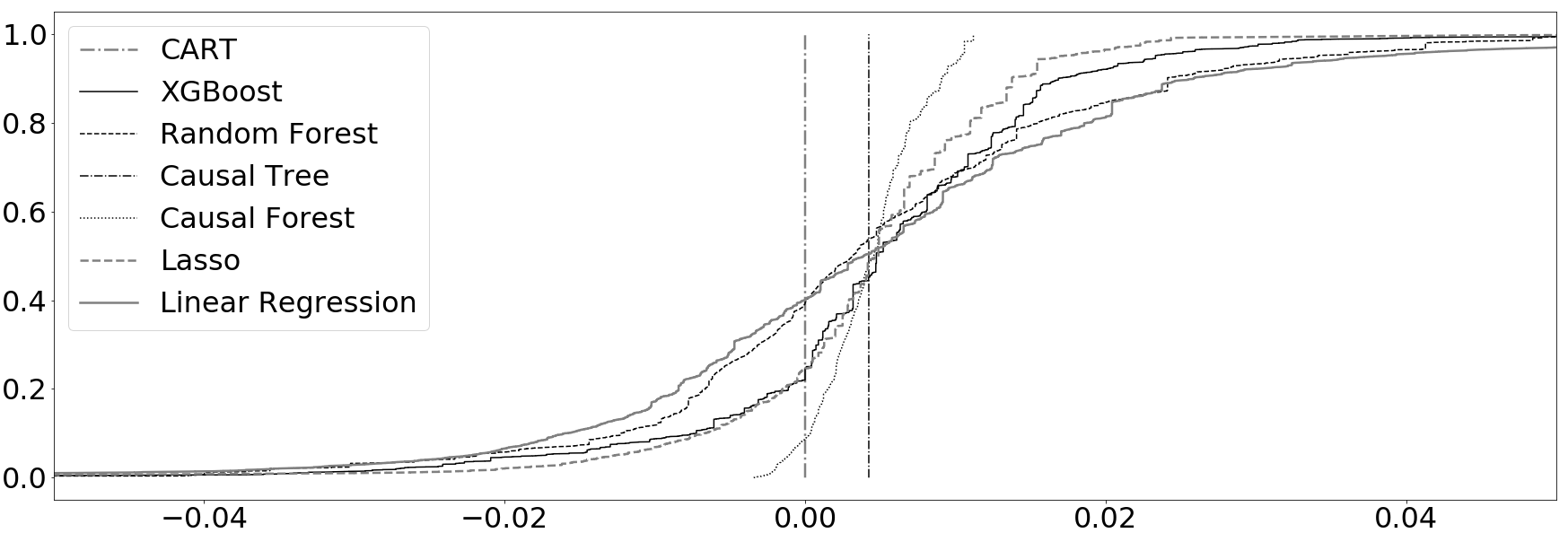}
    CDF of $\mathbf{\tau_{7, 14}}$\par\medskip
    
    \caption{\label{figure:cdf_comparison_all} 
    The CDF of estimated CATEs from using different methods (for test data).} 
\end{figure}

We now examine if there is any relationship between the Conditional Average Treatment Effects (or CATEs) estimated based on a specific method and the performance of the policy based on that method. In Figure \ref{figure:cdf_comparison_all}, we show the CDFs of CATE estimates for the three pairs of treatments ($\tau_{7,30}$, $\tau_{14, 30}$, and $\tau_{7,14}$) based on each of the  methods discussed in $\S$\ref{ssec:estimators}. For expositional purposes, we focus on the top panel in the rest of this section. This panel depicts the CATE estimates for the 7 and 30-day treatments, i.e., $\tau_{7,30}$. The discussions for the middle and bottom panels are similar. 

The first pattern that stands out is that CATE estimates based on CART, causal tree, and casual forest show very little heterogeneity (see the three vertical lines to the right of zero). This explains why policies based on these methods perform poorly -- they are unable to personalize the policy sufficiently to optimize users' response at the individual level. Recall that $\pi_{cart}$ and $\pi_{c\_tree}$ give the same treatment to all users and $\pi_{c\_forest}$ gives the 7-day treatment to over 91\% of users (Table \ref{table:method_length_ratio}). 

In contrast, CATE estimates based on linear regression and random forest show the maximum amount of heterogeneity (see the two rightmost curves). This pattern, in combination with the poor performance of $\pi_{reg}$ and $\pi_{r\_forest}$ on test data (and their extremely good performance on training data) suggests serious overfitting problems. That is, these models seem to infer much more heterogeneity than is true in the data. This, in turn, leads them to personalize policies to a larger extent than is ideal. Recall that $\pi_{reg}$ and $\pi_{r\_forest}$ show the maximum dispersion (or variation) in treatment assignment compared to other policies (Table \ref{table:method_length_ratio}). 

Finally, we see that the CDFs of treatment effects based on lasso and XGBoost lie somewhere in between the above two groups. They show sufficient heterogeneity, but not too much. Hence, policies based on these methods are able to personalize the treatment sufficiently, but also avoid overfitting at the same time. As shown in Table \ref{table:method_length_ratio}, the dispersion in treatment assignment for these two policies is higher than that in $\pi_{cart}$, $\pi_{c\_tree}$, and $\pi_{c\_forest}$, but lower than that in $\pi_{reg}$ and $\pi_{r\_forest}$.

Thus, the ideal estimators for policy design are those that are able to capture sufficient heterogeneity to personalize effectively without overfitting (i.e., capture spurious heterogeneity). 

\subsection{Understanding the Source of these Differences}
\label{ssec:source}
We now examine why some policies perform better than others. First, in $\S$\ref{sssec:quant_gains}, we derive a general mathematical expression to quantify the gains from any policy over the current experiment. Next, in $\S$\ref{sssec:explain_counter}, we use these mathematical expressions to explain the difference in the performance of the various counterfactual policies.

\subsubsection{Quantifying the Gains from a Policy}
\label{sssec:quant_gains}

We now formalize the intuition behind this simple example for a general case. In a fully randomized experiment, the treatment propensities are the same for all observations i.e., $e(W_i = w, X_i) = e_w$. So we can write the IPS reward function for any policy $\pi$ as:
\begin{align}
\notag
     \hat{R}_{IPS}(\pi, Y) &= \frac{1}{N} \sum_{i=1}^N \frac{1[{W_i = \pi(X_i)}] Y_i}{e(W_i, X_i)}\\\notag
     & = \frac{1}{N} \left[\sum_{w \in \mathcal{W}} \left[\sum_{i: \pi(X_i) =W_i=w} \frac{Y_i}{e_w}\right]\right] \\\notag
     & = \frac{1}{N} \sum_{w \in \mathcal{W}}\frac{1}{e_w} \left[\sum_{i:\pi(X_i) =W_i=w} Y_i \right]\\\notag
     & = \frac{1}{N} \sum_{w \in \mathcal{W}} \frac{1}{e_w}  N e_w \upsilon_w \bar{Y}_{[\pi(X_i) =W_i = w]}\\
     \label{eq:IPS_new}
     & = \sum_{w \in \mathcal{W}} \upsilon_w \bar{Y}_{[\pi(X_i) =W_i = w]},
\end{align}
where $\upsilon_w=\sum_{i=1}^N\frac{1[{\pi(X_i)=w}]}{N}$ is the fraction of observations for whom the the policy $\pi$ assigns treatment $w$ and $\bar{Y}_{[\pi(X_i) = W_i = w]}$ is the average outcome for policy congruent observations that are assigned treatment $w$ by policy $\pi$. 

Next, let $\Upsilon(\pi, Y)$ denote the change or improvement in reward from implementing policy $\pi$ compared to the current experiment. We can formally write $\Upsilon(\pi, Y)$ as:
\begin{align}
\notag
     \Upsilon(\pi, Y) &= \hat{R}_{IPS}(\pi, Y) - \frac{1}{N}\sum_{i=1}^N Y_i \\\notag
     & = \sum_{w \in \mathcal{W}}\left[ \upsilon_w \bar{Y}_{[\pi(X_i) = W_i = w]} - \sum_{w' \in \mathcal{W}} \upsilon_w e_{w'} \bar{Y}_{[\pi(X_i) = w, W_i = w']} \right] \\
     & = \sum_{w \in \mathcal{W}} \upsilon_w  \left[ \sum_{w' \in \mathcal{W}} e_{w'} \left(\bar{Y}_{[\pi(X_i) = W_i = w]} -\bar{Y}_{[\pi(X_i) = w, W_i = w']}\right) \right]. \label{eq:diff_explain}
\end{align}
Note that $\frac{1}{N}\sum_{i=1}^N Y_i$ is not a function of the policy $\pi$ being evaluated because it is simply the reward of the randomized experiment currently being implemented. It is therefore constant across different policies. Hence, the policy improvement function $\Upsilon(\pi, Y)$ can be used as a metric to compare the performance of different policies and explain why some policies do better than others.

According to Equation \eqref{eq:diff_explain}, gains from a policy $\pi$ depend on two quantities: (1) For each treatment $w$, the fraction of observations for which that treatment is assigned ($\upsilon_w$), and (2) For these observations, the difference in expected outcomes for policy congruent and policy in-congruent users, i.e., $\left(\bar{Y}_{[\pi(X_i) = W_i = w]} -\bar{Y}_{[\pi(X_i) = w, W_i = w']}\right) \; \forall \; w'$. A good policy is one that ensures that -- (a) the difference between policy congruent and in-congruent users is positive and large for each $w$, and (b) a higher fraction of users ($\upsilon_w$s) get the more positive and higher differences. 

\subsubsection{Explaining the Gains from Counterfactual Policies}
\label{sssec:explain_counter}
We now use the expressions derived above to explain the difference in performances of the counterfactual policies discussed earlier. In Table \ref{table:congruency}, for each policy $\pi$, we show: (1) the fraction of users for whom the policy prescribes the three treatments (7, 14, and 30 days), and (2) the subscription rate (or average outcome) of both policy congruent and in-congruent users (e.g., for  users for whom the policy prescribes 7 days, the subscription rate of all three sets of users -- policy congruent users who received the 7-day treatment, policy in-congruent users who received the 14 and 30 day treatments). Note that the first sub-column within each policy simply shows the fraction of users assigned 7, 14, and 30 days by that policy. It is useful to recognize that these numbers in the bottom panel (for test data) are exactly the same as that in Table \ref{table:method_length_ratio} from $\S$\ref{ssec:policies}.

\begin{sidewaystable}[!htbp]
\centering
\footnotesize
\begin{center}
    \textbf{Training Dataset}
\end{center}
\vspace{-0.4cm}
\begin{tabular}{cc|cc|cc|cc|cc|cc|cc}
\toprule
\multicolumn{2}{c}{\textbf{Trial Length }}& 
\multicolumn{2}{c}{$\pi_{7}$}& 
\multicolumn{2}{c}{$\pi_{reg}$}&
\multicolumn{2}{c}{$\pi_{lasso}$}&
\multicolumn{2}{c}{$\pi_{r\_forest}$}&
\multicolumn{2}{c}{$\pi_{xgboost}$}&
\multicolumn{2}{c}{$\pi_{c\_forest}$}\\\hline
\textbf{Optimal} & \textbf{Actual}&  \textbf{Frac.} & \textbf{Sub. Rate}&\textbf{Frac.} & \textbf{Sub. Rate}&\textbf{Frac.} & \textbf{Sub. Rate}&\textbf{Frac.} & \textbf{Sub. Rate}&\textbf{Frac.} & \textbf{Sub. Rate}&\textbf{Frac.} & \textbf{Sub. Rate} \\
\hline\hline
\multirow{3}{*}{7 Days} & 7 Days &   
\multirow{3}{*}{1}  &  \cellcolor[gray]{0.8}15.320 & 
\multirow{3}{*}{0.522}  &  \cellcolor[gray]{0.8}14.224 &  \multirow{3}{*}{0.688}  &  \cellcolor[gray]{0.8}10.686 &  \multirow{3}{*}{0.445}  &  \cellcolor[gray]{0.8}14.170 &  \multirow{3}{*}{0.698}  &  \cellcolor[gray]{0.8}13.409 &  \multirow{3}{*}{0.913}  &  \cellcolor[gray]{0.8}15.143 \\
& 14 Days &  & 14.897 & &  12.360 &  &   9.451 &   &  10.360 &  &  11.969 &    &  14.388\\
 & 30 Days & & 14.684 & &   12.569 & &   9.575 & &  10.404 & &  12.153 &  &  14.364\\
\midrule
\multirow{3}{*}{14 Days} & 7 Days & 
\multirow{3}{*}{---} & --- &
\multirow{3}{*}{0.322} &  18.878 &  \multirow{3}{*}{0.233} &  28.449 &  \multirow{3}{*}{0.271} &  19.897 &  \multirow{3}{*}{0.184} &  22.268 &  \multirow{3}{*}{0.087} &  17.146\\

& 14 Days & & ---& &  \cellcolor[gray]{0.8}20.549 &  &  \cellcolor[gray]{0.8}30.288 &  &  \cellcolor[gray]{0.8}24.139 &  &  \cellcolor[gray]{0.8}25.209 &   &  \cellcolor[gray]{0.8}20.158\\
 & 30 Days &  &---& &  19.503 &  &  28.547 &  &  20.228 &  &  22.862 &  &  18.084\\
\bottomrule
\end{tabular}
\begin{center}
    \textbf{Test Dataset}
    \vspace{-0.25cm}
\end{center}
\begin{tabular}{cc|cc|cc|cc|cc|cc|cc}
\toprule
\textbf{Optimal} & \textbf{Actual}&  \textbf{Frac.} & \textbf{Sub. Rate}&\textbf{Frac.} & \textbf{Sub. Rate}&\textbf{Frac.} & \textbf{Sub. Rate}&\textbf{Frac.} & \textbf{Sub. Rate}&\textbf{Frac.} & \textbf{Sub. Rate}&\textbf{Frac.} & \textbf{Sub. Rate} \\
\hline\hline
\multirow{3}{*}{7 Days} & 7 Days & 
\multirow{3}{*}{1}  &  \cellcolor[gray]{0.8}15.445 & 
\multirow{3}{*}{0.521}  &  \cellcolor[gray]{0.8}13.406 &  \multirow{3}{*}{0.689}  &  \cellcolor[gray]{0.8}10.538 &  \multirow{3}{*}{0.444}  &  \cellcolor[gray]{0.8}11.401 &  \multirow{3}{*}{0.697}  &  \cellcolor[gray]{0.8}13.281 &  \multirow{3}{*}{0.911}  &  \cellcolor[gray]{0.8}15.145 \\
& 14 Days &  & 15.107 &  &  12.924 & &   9.741 &  &  11.407 & &  12.786 & &  14.764\\
 & 30 Days & & 14.627 & &  12.515 & &   9.621 & &  10.865 & &  12.122 & & 14.252\\
\midrule
\multirow{3}{*}{14 Days} & 7 Days & 
\multirow{3}{*}{---} & --- &
\multirow{3}{*}{0.322} &  20.499 &  \multirow{3}{*}{0.232} &  29.720 & 
\multirow{3}{*}{0.269} &  22.447 &  \multirow{3}{*}{0.185} &  23.868 &  \multirow{3}{*}{0.089} &  18.524\\

& 14 Days & &  --- &  & \cellcolor[gray]{0.8}20.193 & & \cellcolor[gray]{0.8}30.316 & & \cellcolor[gray]{0.8}20.452 & & \cellcolor[gray]{0.8}23.570 & & \cellcolor[gray]{0.8}18.707\\
 & 30 Days & &  --- & &  19.437 & &  28.710 & &  20.630 & &  23.036 & &  18.447 \\
 \midrule
 \multirow{3}{*}{30 Days} & 7 Days & 
 \multirow{3}{*}{---} & --- &
 \multirow{3}{*}{0.157} &  11.809 &  \multirow{3}{*}{0.079} &  17.172 &  \multirow{3}{*}{0.287} &  15.189 &  \multirow{3}{*}{0.118} & 15.037  &   
 \multirow{3}{*}{---} &  --- \\
\bottomrule
\end{tabular}
\caption{\label{table:congruency} For each counterfactual policy, we show the fraction of users that the policy assigns to the 7, 14, and 30 day treatments, the subscription rate of policy congruent users (grey cells) and in-congruent users (white cells), and $\Upsilon(\pi, Y)$. For example, $\Upsilon(\pi_{lasso}, Y)$ on test data is calculated as follows: $\Upsilon(\pi_{lasso}, Y) = 
    0.689 * \left[0.152*0 + 0.149*0.797 +  0.699*0.917 \right] 
      + 0.232 * \left[0.149*0.596 + 0.150*0 + 0.700*1.606 \right]
     + 0.079 * \left[0.148* -0.257 + 0.153*-0.119 + 0.698*0 \right] = 0.800$. The top panel is on training data and bottom panel on test data.} 
\end{sidewaystable}

We now discuss the performance of the different counterfactual policies in the training data (top panel of Table \ref{table:congruency}). First, consider $\pi_{reg}$. Here, the difference between policy congruent and policy in-congruent cells for each treatment is always positive, e.g., for the set of users for whom the policy assigns 7 days, the average outcome for policy congruent users is $14.224$ (grey cell), which is greater than both $12.36$ and $12.569$ (the white cells). The same pattern is true for the rest of the policies in the training data -- for any treatment $w$, the average outcome of policy congruent users is always greater than that of policy in-congruent users. 

However, things look very different when we look at the performance of these policies in the test data (bottom panel). In the case of $\pi_{reg}$, we see that the average outcome for policy congruent users is not always the highest. For example, the average outcome for policy congruent users assigned the 30-day trial is $11.761$. However, if these users had instead received the 7 or 14-day trials (which the policy deemed sub-optimal for them), they would have done better (with average subscription rates of $11.809$ and $11.993$, respectively)! $\pi_{reg}$ is similarly sub-optimal for users for whom it prescribes the 14-day treatment. Together, these two sets of users constitute $47.9\%$ ($0.157 + 0.322 = 0.479$) of the population. Thus, the policy $\pi_{reg}$ is sub-optimal for a large fraction of users and the extent of sub-optimality (the differences in the outcomes for policy congruent and in-congruent users) is high. Hence, the overall performance of this policy is poor.

In contrast, if we look at $\pi_{lasso}$, we see that for users whom it assigns to the 7 and 14-day treatments, the average outcome for the policy congruent cells is higher than that of policy in-congruent cells. More importantly, these differences are substantial and happen for a large fraction of users ($0.689 + 0.232 = 0.921$). We do find that this policy is somewhat sub-optimal for the users it assigns to the 30-day treatment ($16.914$ is less than both $17.172$ and $17.033$). However, these differences are quite small ($16.914 - 17.172 \approx -0.26$ and $16.914 - 17.033 \approx 0.12$) and the fraction of users for whom it makes this mistake is also small ($0.079$). Overall, this is the reason behind the superior performance of $\pi_{lasso}$: it is able to improve the response of a vast majority of users (over 92\%) by a substantial amount; and when it does make a mistake in treatment allocation, it is for a small set of users and the magnitude of this mistake is quite small. 

Next, we can see why $\pi_{xgboost}$ is the second-most effective policy after $\pi_{lasso}$: it is sub-optimal for users whom it prescribes the 14-day treatment. While the difference between the expected outcomes for policy congruent and in-congruent users is not very high, the fraction of users for whom this mistake happens is higher than that in the case of lasso ($18.5\%$).

It is also apparent why the policy based on random forest, $\pi_{r\_forest}$, performs so poorly -- it assigns sub-optimal treatments to all the users. Notice that in no condition (or treatment) is the grey cell the highest for that condition. Finally, we look at the policy based on causal forest, $\pi_{c\_forest}$. Interestingly, this is the only policy that does not make sub-optimal allocations. For both sets of users (those assigned to the 7-day and 14-day treatments), the policy congruent cells are always highest. However, the problem with this policy is that these differences are minor (e.g., in the case of users whom it assigns to the 7-day treatment, the gain of going from 14-days to 7-days is $0.381$). Thus, $\pi_{c\_forest}$ is not able to sufficiently personalize treatments. It is thus similar to the $\pi_7$, which also does not make sub-optimal allocations (but also does not personalize treatment assignment).

\section{Substantive Findings: Segmentation, Customer Loyalty, and Revenues}
\label{sec:sub_finding}
In this section, we conduct a series analysis based on the best personalized policy ($\pi_{lasso}$) and derive a set of substantive findings of relevance to SaaS firms in the free trial context. In $\S$\ref{ssec:segmentation}, we present some descriptive analysis to explain why/how free trial assignment varies across consumers under the optimal personalized policy. Then, in $\S$\ref{ssec:revenue}, we discuss the improvements in long-term outcomes like customer loyalty and revenue improvements under different counterfactual free trial policies. 

\subsection{Segmentation Based on Policy-prescribed Treatment}
\label{ssec:segmentation}

So far, we focused on the question of ``Who (should get a treatment)''. We now examine the question of ``Why (should s/he get a specific treatment)''. Understanding why some users respond well to longer trials while others respond better to shorter trials can give us insight into consumers' preferences and decision-making process. These insights are valuable for two reasons. First, from the firm's perspective, they can be leveraged to improve other marketing interventions such as advertising and pricing. Second, from a research perspective, this gives us a better understanding of the heterogeneity in the effectiveness of trial length on subscription, which can be extrapolated to other firms and settings.

We now segment consumers based on their prescribed treatment (according to $\pi_{lasso}$, since it is the best policy) and correlate their prescribed treatment with pre-treatment demographic and post-treatment behavioral variables. Since these variables are not randomized across individuals, we cannot make any causal inference on their effect on users' responsiveness. Rather, the goal is to present some suggestive evidence for the underlying mechanisms driving users' behavior. 

In Appendix \ref{sec:app_segment}, we present Tables \ref{table:correlate_pre} and Table \ref{table:correlate_post}, that show the distributions of pre-treatment demographic variables and post-treatment behaviors for three segments:
\squishlist
\item Segment 1: Users who are prescribed the 30-day treatment, i.e., $\pi_{lasso}(X_i) = 30$ days.
\item Segment 2: Users who are prescribed the 14-day treatment, i.e., $\pi_{lasso}(X_i) = 14$ days.
\item Segment 3: Users who are prescribed the 7-day treatment, i.e., $\pi_{lasso}(X_i) = 7$ days.
\squishend

We now discuss the attributes and behavior of these three segments. 
\squishlist
\item {\bf Segment 1:} These individuals are more likely to be {\it legacy} users. They are more experienced than average, are less likely to be students and hobbyists, and more likely to sign up through the app manager instead of the website. All these factors suggest that these are more likely to be legacy users who are already aware of and using some version of the software. This is also corroborated by their post-treatment behavior -- conditional on subscription, e.g., the firm is able to retain them a bit longer than average. 

\item {\bf Segment 2:} These individuals are more likely to be {\it learners}. They are more likely to be beginners or mixed skill users, and have the highest usage and subscription rate. Thus, these users benefit the most from the free trial since they seem to actually try the product's features and evaluate the product the most before deciding whether to subscribe or not.\footnote{An interesting question that arises here is this: if these users use the product and learn the software during the free trial period, why is the longer 30-day free trial not optimal for them? There are two possible explanations here. First, with a longer trial, these users might decide to free-ride and use the product for their needs, but not subscribe. Second, it is possible that the features that users learn about later in a free trial are less appealing and have a negative impact on subscription \citep{Heiman_Muller_1996}.}

\item {\bf Segment 3:} These individuals are more likely to be {\it free-riders}. A vast majority of them are beginners or students and are the least likely to subscribe. These users are likely to be individuals who need the software for specific tasks, such as course projects. As a result, they are more likely to free-ride if given a lengthy free trial. A short free trial therefore leads to a higher subscription rate in this group.

\squishend
It worth mentioning that our findings disprove some of the previous theories on the relationship between users' experience and the effectiveness of free trials. For example, \cite{Dey_etal_2013} argue that beginners should be given longer free trials because longer trials allow them to learn about the product, which increases their likelihood of subscription. However, we find the opposite -- shorter trials are more effective for beginners and longer trials are better for experienced users. This is because more experienced users are likely to be familiar with the functionalities in the legacy software and need more time to evaluate these functionalities in the new cloud service and compare it with the legacy version. On the other hand, beginners are just starting to get to know the software, and are unlikely to have enough information to evaluate the software. Given that this firm is the dominant player in this market and sets the standards in this industry, a beginner's decision to subscribe or not is likely driven by the brand name rather than their experience during the free trial. 

\subsection{Long-term Consumer Loyalty and Profitability}
\label{ssec:revenue}
So far we have focused on designing and evaluating policies that maximize subscriptions. However, a policy that maximizes subscriptions (or short-run conversions) may not be the best long-run policy if it brings in users who are less ``profitable'' or less ``loyal''. For example, a policy that increases subscriptions among students (who get a significant educational discount and hence pay lower prices) and/or users who subscribe to lower-end products/bundles (that are priced much lower than the all-inclusive software suite) at the expense of high-end users can lead to lower revenues. Similarly, some policies may increase subscriptions, but do so at the expense of long-term retention, i.e., they may bring in the less loyal consumers who churn within a short period. Thus, a subscription-optimal policy may in fact be sub-optimal from the perspective of long-run outcomes \citep{Gupta_etal_2006, Fader_Hardie_2009, Mccarthy_etal_2017}. 

In this section, we therefore examine the performance of the policies discussed earlier on two important post-subscription outcomes of interest for the firm.
\begin{enumerate}
    \item Consumer loyalty, as measured by subscription length or the number of months a user subscribes to the service over the two year period after the experiment.
    \item Consumer profitability, as measured by the revenue generated by the user over the two years after the experiment. (In SaaS settings, revenues and profits can be treated as equivalent since the marginal cost of serving an additional user is close to zero.) 
\end{enumerate}
Recall that the summary statistics of subscription length and revenues were shown in Table \ref{table:post_summary}.

We first describe the procedure for deriving the IPS estimates of the average subscription length and revenue under a given policy. For each policy $\pi$:
\squishlist
\item We first segment users into three groups based on the policy-prescribed treatment: (1) $\pi(X_i) = 7$ days, (2) $\pi(X_i) = 14$ days, and (3)  $\pi(X_i) = 30$ days.  
\item Then, to obtain IPS estimate of average subscription length and revenue under policy $\pi$, we use Equation \eqref{eq:ips_reward_emp} and the observed subscription lengths and revenues as the outcome variable ($Y_i$). One minor point is that we do not have access to revenue data for all subscribers (recall the discussion in $\S$\ref{sssec:post_data}). So the revenue calculations are done on the subset of users for whom we have non-missing revenue data.
\squishend

\begin{table}[t]
\centering
\footnotesize

\begin{tabular}{ll|cc|cc|cc|cc}
\toprule
\multirow{3}{*}{\textbf{Category}} & \multirow{3}{*}{\textbf{Policy}}&
\multicolumn{4}{c}{\textbf{Subscription Length}} & 
\multicolumn{4}{c}{\textbf{Revenue}} \\
& & \multicolumn{2}{c}{\textbf{Estimate (Months)}}&
\multicolumn{2}{c}{\textbf{Increase (\%)}}& \multicolumn{2}{c}{\textbf{Estimate (\$)}}&
\multicolumn{2}{c}{\textbf{Increase (\%)}}\\
& & Training & Test & Training & Test & Training & Test & Training & Test
\\
\hline\hline
\multirow{3}{*}{\shortstack[l]{One length \\ for all}} &$\pi_{30}$ (Baseline) & 2.17 & 2.19 & --- & --- & 77.31 & 78.41 & --- & ---  \\
& $\pi_{14}$& 2.22 & 2.32 & 2.59 & 5.91 & 79.86 & 84.85 & 3.30 & 8.21 \\
& $\pi_{7}$ & 2.32 & 2.33 & 7.27 & 6.28 & 83.51 & 84.52 & 8.03 & 7.80 \\
\midrule
\multirow{5}{*}{\shortstack[l]{Based on \\ outcome \\estimation}}& $\pi_{reg}$   &  2.39 &      2.31 &           10.19 &           5.46 &      86.73 &     84.47 &           12.19 &           7.73 \\
& $\pi_{lasso}$  & 2.39 & 2.36 & 10.42 & 7.96 & 86.90 & 87.51 & 12.41 & 11.61\\
& $\pi_{cart}$  &  2.32 & 2.33 & 7.27 & 6.28 & 83.51 & 84.52 & 8.03 & 7.80 \\
& $\pi_{r\_forest}$ & 2.62 &      2.25 &           20.91 &           3.05 &      94.15 &     80.50 &           21.79 &           2.67 \\
& $\pi_{xgboost}$  & 2.41 & 2.36 & 11.42 & 7.85 & 87.76 & 86.72 & 13.52 & 10.60 \\
\midrule
\multirow{2}{*}{\shortstack[l]{Based on treatment \\ effect estimation}} & $\pi_{c\_forest}$ & 2.38 & 2.32 & 9.92 & 5.98 & 85.48 & 84.08 & 10.57 & 7.24 \\
& $\pi_{c\_tree}$ &2.32 &      2.33 &            7.27 &           6.28 &      83.51 &     84.52 &            8.03 &           7.80 \\
\bottomrule
\end{tabular}
\caption{\small\label{table:method_comparison_rev} IPS estimates of the average subscription length and revenue under different counterfactual policies. All the revenue numbers are scaled by a constant factor to preserve the firm's anonymity. The number of observations used for revenue calculation is  $234,237$ in the training data and $99,986$ in the test data. The corresponding numbers for subscription length are similar to Table \ref{table:data_distribution}}.
\end{table}

The IPS estimates of subscription length and revenues for the counterfactual policies are shown in Table \ref{table:method_comparison_rev}. (By definition, these IPS estimates are for the full population, and not just subscribers.) The main takeaway is that $\pi_{lasso}$ performs the best here also, followed by $\pi_{xgboost}$.\footnote{Tables \ref{table:paired_t_test_month} and \ref{table:paired_t_test_rev} in Appendix $\S$\ref{sec:app_ttest} show the results for paired t-tests comparing the gains in subscription length and revenue across policies.} Thus, there are no concerns about this policy bringing in more subscriptions at the expense of long-run consumer loyalty or revenue.\footnote{Note that $\pi_{14}$ outperforms $\pi_7$ in the test data (on revenue) even though $\pi_7$ is the best policy in the training data. We present a brief explanation for this discrepancy now. In general, estimates from one data set are valid in another data set only when the joint distribution of outcomes and covariates are similar in both data sets. This is usually true when the data set is very large or the signal-to-noise ratio is low. However, as discussed in $\S$\ref{ssec:datasets}, neither of these are true in our case; so there are some minor differences in our training and test data due to the randomness in the splitting procedure. The main difference is this -- the distributions of subscription length for the 14-day condition in the training and test data are different. This is however not the case for the 7- or 30-day conditions; see Table \ref{table:retention_summary} in Appendix \ref{sec:app_gain_sub_rev}. Thus, the estimate of subscription length from training data does not translate well to test data, and this leads to the large difference in the subscription length and revenue estimates across the training and test data sets.}

An interesting empirical pattern here is that the gains in subscription length and revenues are quantitatively different from the gain in subscription (comparing the percentage increases in Tables \ref{table:method_comparison_sub} and \ref{table:method_comparison_rev}). We now discuss the source of this difference. 

We can expand the expected subscription length (denoted by $Y_i^l$) conditional on treatment $W_i$ for the population of users as follows:
\begin{equation}
\E(Y^l_i|W_i) = \textrm{Pr}(Y^s_i|W_i) \cdot \E \left[T_{end} - T_{start}|W_i, Y^s_i=1\right],
\label{eq:sub_expansion}
\end{equation}
where $\textrm{Pr}(Y^s_i|W_i)$ is the probability that user $i$ will subscribe conditional on receiving treatment $W_i$ and $\E \left[T_{end} - T_{start}|W_i, Y^s_i=1\right]$ is $i$'s expected length of subscription conditional on receiving treatment $W_i$ and subscribing ($Y_i^s = 1$). The reason for the discrepancy in the gains on the two outcomes -- subscription and subscription length -- becomes apparent from Equation \eqref{eq:sub_expansion}. If trial length affects not just subscription, but also how long a subscriber will remain loyal to the firm, then the gains in $Y_i^l$ will be naturally different from the gains in subscription. To examine if this is true in our data, we show the summary statistics for $\E \left[T_{end} - T_{start}|W_i, Y^s_i=1\right]$ for the three trial lengths in Table \ref{table:retention_summary} in Appendix \ref{sec:app_gain_sub_rev}. We see that there are some small differences in this metric across the three trial lengths, which account for the differences between the gains in subscription and gains in subscription length.

Similarly, we can write the expected revenue (denoted by $Y_i^r$) conditional on treatment $W_i$ for the population as:
\begin{equation}
\E(Y^r_i|W_i) = \textrm{Pr}(Y^s_i|W_i) \cdot \E \left[T_{end} - T_{start}|W_i, Y^s_i=1\right] \cdot \E \left[\textrm{Price}_i|W_i, X_i, Y^s_i=1\right].
\label{eq:revenue_expansion}
\end{equation}
This is similar to Equation \eqref{eq:sub_expansion}, with the additional $\E \left[\textrm{Price}_i|W_i, X_i, Y^s_i=1\right]$ term. It suggests that trial length can influence revenues through three channels -- (1) subscriptions, (2) length of subscription, and (3) price of the product subscribed. The first two were already discussed in the paragraph above. We now examine whether the products that consumers subscribe to and the prices that they pay are also a function of trial length. That is, we examine whether $\E \left[\textrm{Price}_i|W_i, X_i, Y^s_i=1\right]$ is indeed a function of $W_i$ in our data. Note that the price that a subscriber pays is a function of both the product that s/he subscribes to (e.g., single product, all-inclusive bundle) as well her/his demographics (e.g., students pay lower prices for the same product). In Table \ref{table:subscription_summary} in Appendix \ref{sec:app_gain_sub_rev}, we present the distribution of products and subscription type (educational, commercial, or government) by trial length for all the subscribers in our data. Again, we see that there are some minor differences in product and subscription types across trial lengths. However, the main reason why revenue gains are significantly different from subscription gains is likely due to the difference in the samples used to calculate the two gains. Recall that revenue gains are calculated on a sample of $334,223$ users for whom revenue data are available, whereas subscription gains are calculated on the full sample of $337,724$ users. 

In sum, there are two main takeaways from this section. First, personalized policies designed to optimize short-term outcomes (such as conversion/subscription) also perform well on long-term outcomes (at least in our setting). Second, the magnitude of gains can be different across outcomes since the treatment can affect long-run outcomes through multiple channels (in addition to conversion). In our setting, we find that trial length only plays a marginal role on other factors that affect revenue or loyalty; so the magnitude of the gains across outcomes are similar (controlling for the sample). However, these effects can vary across settings and policy-makers should keep them in mind when designing and evaluating personalized policies.

\section{Conclusions}
\label{sec:conclusion}
Free trials promotions are now a commonly used acquisition strategy in the SaaS industry. In this paper, we examine -- (1) how a firm can personalize the length of free trials based on a customer's observed demographics or skill level, and (2) document the returns to personalized free trials on both short-term conversions and long-run outcomes such as customer loyalty and revenues.

We first propose a general framework for personalized targeting policy design and evaluation that is compatible with a high-dimensional covariate space. Our framework consists of three components and is applicable to any setting where researchers have data from a well-designed experiment that satisfies the two important properties of unconfoundedness and compliance. In the first component, we define the optimal policy design problem and present a two-step solution concept. In the first stage, we learn flexible functions of either outcomes or pairwise heterogeneous treatment effects. Then, in the second stage, we use the function(s) learned in the first stage to assign the optimal treatment for each user. In the second component, we design personalized policies based on five outcome estimators (linear regression, lasso, CART, random forest, and boosted regression trees) and two heterogeneous treatment effects estimators (causal tree and causal forest). In the third component, we use the Inverse Propensity Score (IPS) reward estimator to evaluate the reward or gain from any targeting policy.

We apply our framework to data from a large-scale field experiment on free trials conducted by a leading SaaS firm. The firm manipulated the length of free trials such that new users were randomly assigned to 7, 14, or 30 days of free trial. In this setting, the 7-day trial improves subscription by $5.59\%$ over the baseline of 30 days for all policy. We also find that the two outcome estimators -- lasso and XGBoost -- perform the best in our setting, i.e., maximize conversions. In contrast, the two heterogeneous treatment effects estimators -- causal tree and causal forest -- perform poorly because they are unable to personalize the policy sufficiently. Indeed, a key takeaway of our research is that while personalization can lead to substantial benefits, many commonly used methods to personalize are often ineffective. Hence, researchers and managers need to use appropriate policy design and evaluation methods for effective personalization. Finally, we show that, personalized free trial policies designed to maximize short-run conversions also perform well on long-run outcomes such as consumer loyalty and profitability in SaaS settings. We expect our substantive findings to be of relevance to SaaS firms and managers who want to optimize their free trial promotions. 

Today's technology and digital marketing firms have unprecedented access to individual-level data on consumers' attributes, the ability to experiment at scale, and customize marketing actions in real-time. This has led to significant interest in the question of how to personalize marketing interventions. However, so far researchers and practitioners have little theoretical or empirical guidance in this task. We hope that the general framework presented in the paper combined with the empirical lessons comparing the performance of a series of estimators will help managers to effectively personalize marketing-mix variables in other settings.

\newpage
\renewcommand{\baselinestretch}{1.0} \small{
  \bibliographystyle{abbrvnat} \bibliography{ref} }

\newpage

\begin{appendices}
\setcounter{table}{0}
\setcounter{page}{1}
\setcounter{figure}{0}
\setcounter{equation}{0}
\renewcommand{\thetable}{A\arabic{table}}
\renewcommand{\thefigure}{A\arabic{figure}}
\renewcommand{\theequation}{A.\arabic{equation}}
\renewcommand{\thepage}{\roman{page}}

\section{Heterogeneous Treatment Effects Estimators}
\label{sec:app_het_treat}
We present additional details of the causal tree and generalized random forest models that we use for estimating heterogeneous treatment effects.

\subsection{Causal Tree}
\label{ssec:app_causal_tree}
The main idea behind causal tree is that we can use recursive partitioning to estimate heterogeneous treatment effects if we can come up with an objective function that can identify partitions with similar within-partition treatment effect.\footnote{The first example of such a criterion was proposed by \citet{Hansotia_Rukstales_2002}: for each potential split, the algorithm calculates the difference between the average outcome of the treatment and control in the right ($\Delta Y_r$) and left ($\Delta Y_l$) branches of the tree. Then, it selects the split with the highest value of $\Delta \Delta Y = |\Delta Y_r - \Delta Y_l|$. Other splitting criteria include maximizing the difference between the class distributions in the treatment and control groups \citep{Rzepakowski_Jaroszewicz_2012}.} \citet{Athey_Imbens_2016} show that maximizing the variation in the estimated treatment effects achieves this objective, which can be written as:
\begin{equation}
    Var[\hat{\tau}(X)] =  \frac{1}{N}\sum_{i=1}^N 
    \hat{\tau}^2(X_i) - \left(\frac{1}{N}\sum_{i=1}^N
    \hat{\tau}(X_i)\right)^2
    \label{eq:causal_tree_objective}
\end{equation}
Since $\left(\frac{1}{N}\sum_{i=1}^N \hat{\tau}(X_i)\right)^2$ remains constant with respect to any possible next split, the objective function can be also described as maximizing the average of the square of estimated treatment effects. In practice, the algorithm chooses the split that maximizes $Var[\hat{\tau}(X)] - \zeta T$, where the second term is added for complexity control and is analogous to the regularization term in CART. 

The algorithm consists of two steps. In the first step, it recursively splits the covariate space into partitions. In the second step, it estimates the treatment effect within each partition ($l(x)$) using Equation \eqref{eq:potential_outcome_lx}. Intuitively, this algorithm pools observations with similar treatment effects into the same partition because splitting observations that have similar treatment effects does not increase the objective function (i.e., variation in the post-split treatment effect). 

\subsubsection{Generalized Random Forest}
\label{ssec:app_grf}

The Generalized Random Forest (GRF) algorithm takes intuition from causal tree and combines it with ideas from predictive random forests to learn problem-specific kernels for estimating any quantity of interest at a given point in the covariate space. For the estimation of heterogeneous treatment effects, the method proceeds in two steps.

In the first step, it builds trees whose objective is to increase the variation in the estimated treatment effects. Each tree is built on a random sub-sample of the data and random sub-sample of covariates. At each step of the partitioning process (when building a tree), the algorithm first estimates the treatment effects in each parent leaf $P$ by minimizing the R-learner objective function. This objective function is motivated by Robinson's method \citep{Robinson_1988, Nie_Wager_2017} and can be written as follows:
\begin{equation}
    \hat{\tau}_P(\cdot) =  \argmin\limits_{\tau} \left[ \frac{1}{n_P} \sum_{i=1}^{n_P} \Big (\big(Y_i-\hat{m}^{(-i)}(X_i)\big)-(W_i - \hat{e}^{(-i)}(X_i))\hat{\tau}(X_i)\Big)^2 + \Lambda_n(\tau(.))\right],
    \label{eq:r-learner}
\end{equation}
where $n_P$ refers to the number of observations in the parent partition,  $\Lambda_n(\tau(.))$ is a regularizer that determines the complexity of the model used to estimate $\tau_P(.)$s, and  $\hat{m}^{(-i)}(X_i)$ and $\hat{e}^{(-i)}(X_i)$ are out-of-bag estimates for the outcome and propensity score, respectively. While any method can be used for estimating $\hat{m}^{(-i)}(X_i)$ and $\hat{e}^{(-i)}(X_i)$, the GRF implementation uses random forests to estimate these values.

Then, it chooses the next split such that it maximizes the following objective function:
\begin{equation}
\frac{n_L \cdot n_R}{(n_P)^2} \left(\hat{\tau}_L - \hat{\tau}_R \right)^2,
\end{equation}
where $n_L$ and $n_R$ refer to the number of observations in the post-split left and right partitions, respectively. However, instead of calculating the exact values of $\hat{\tau}_L$ and $\hat{\tau}_R$ for each possible split (and for each possible parent), the causal forest algorithm uses a gradient-based approximation of $\hat{\tau}$ for each child node to improve compute speed; see \citet{Athey_etal_2019} for details. Thus, at the end of the first step, the method calculates weights, $\alpha_i(x)$, that denote the frequency with which the $i$-th training sample falls into the same leaf as $x$ in the first step. Formally, $\alpha_i(x)$ is given by $\alpha_i(x) = \frac{1}{B}\sum_{b=1}^B \alpha_{bi}(x)$, where $\alpha_{bi}(x) = \frac{\textbf{1}({X_i \in l_b(x)})}{|l_b(x)|}$, $B$ is the total number of trees built in the first step, and $l_b(x)$ is the partition that $x$ belongs to in the $b$-th tree. 

In the second step, the algorithm uses the idea of a weighted kernel regression to calculate the treatment effect at each point $x$ using weights $\alpha_i(x)$ as follows:
\begin{equation}
    \hat{\tau}(x) =  \frac{ \sum\limits_{i = l}^{N} \alpha_i(x) \left(Y_i - \hat{m}^{(-i)}(X_i)\right)\left(W_i - \hat{e}^{(-i)}(X_i)\right)} {\sum\limits_{i = l}^{N} \alpha_i(x) \left(W_i - \hat{e}^{(-i)}(X_i)\right)^2},
\label{eq:cate_cf}
\end{equation}
As with all supervised learning models, we need to do hyper-parameter optimization to prevent causal forest from overfitting. We refer readers to Appendix $\S$\ref{sec:app_hyperopt} for details on this.

\section{Hyper-parameter Optimization for the Models Estimated}
\label{sec:app_hyperopt}

For each model that we use, we describe the hyper-parameters associated with it and the optimal hyper-parameters that we derive after tuning. In all cases, we use five-fold cross-validation to optimize the hyper-parameters. We then train a model on the entire training data using the optimal hyper-parameters, and report the performance of this model on both the training and test data. 

\squishlist 
\item Least squares does not use any hyper-parameters and hence does not require validation. In this case, we simply train the model on the full training data to infer the model parameters and report the model's performance on both the training and test data.

\item For lasso, the validation procedure is straightforward. The only hyper-parameter to tune is the L1 regularization parameter, $\lambda$. We use the standard cross-validation procedure implemented in the {\it glmnet} package in R. It searches over 98 different values of $\lambda$ ranging from $1.8 \times 10^{-5}$ to $1.5 \times 10^{-1}$, and picks the one that gives us the best out-of-sample performance (based on cross-validation). In our case, it $\lambda=3.1 \times 10^{-4}$. 

\item For CART, we use the package \textit{rpart} in R, which implements a single tree proposed by \citep{breiman_etal_1984}. We only need to pick the complexity parameter ($\zeta$) using cross validation in this case. We search over 3886 different values for $\zeta$ ranging from $8.6 \times 10^{-11}$ to $1.7 \times 10^{-1}$, and derive the optimal complexity parameter as $5.4 \time 10^{-5}$. 

\item For Random Forest, we use the package \textit{sklearn} in Python. There are three hyper-parameters in this case -- (1) $n_{tree}$, the number of trees over which we build our ensemble forest, (2) $\max_f$, the maximum number of features the algorithm try for any split (it can be either all the features or the squared root of the number of features), and (3) $n_{\min}$, the minimum number of samples required to split an internal node. 

The standard method for finding hyper-parameters is grid-search. However, grid-search is very costly and time-consuming when we have to tune many hyper-parameters. So we use the hyperopt package for tuning the hyper-parameters in this model. Hyperopt provides an automated and fast hyper-parameter optimization procedure that is less sensitive to researcher's choice of searched hyper-parameter values; see \citet{Bergstra_etal_2011, Bergstra_etal_2013} for details.

For each of these hyper-parameters, we now define the range over which we search as well as the optimal value of the hyper-parameter are shown below:
\squishlist
\item $n_{tree} \in [100,1200]$ and $n_{tree}^*=1000$
\item $\max_f \in \{n, sqrt(n)\}$ and $\max_f^*=n$
\item $n_{\min} \in [10,300]$ and $n_{\min}^*=70$
\squishend

\item XGBoost also has many hyper-parameters that need tuning. However, we found that our results is sensitive to only three of the parameters: $\alpha$, $\eta$, and $d_{\max}$. The first parameter, $\alpha$, is an L1 regularization parameter, $\eta$ is the shrinkage parameter or learning rate, $d_{\max}$ is maximum depth of trees. Again, we use the hyperopt package to search over a wide range of parameter values. The optimal values are shown below:
\squishlist
\item $\alpha \in \{0.1,0.2,0.5,1,2,5,10,15,20,25\}$ and $\alpha^* = 20$
\item $\eta \in [0,1]$ and $\eta^* = 0.59$
\item $d_{\max} \in \{6,8,10,12\}$ and $d_{\max}^* = 12$
\squishend

\item Causal tree has two hyper-parameters that needs tuning -- (1) the complexity parameter ($\zeta$) and the minimum number of treatment and control observations in each leaf ($q$). We use the cross-validation procedure in the "causalTree" package in R for tuning $\zeta$. We manually tune $q$ using grid-search over the range $[100,1000]$ in increments of $100$. We search over all possible values of $\zeta$ for each $q$. 

The optimal hyper-parameters for the three trees (one for each pair of treatments) are:
\squishlist
\item The tree for 7 and 14 days pair: $\zeta = 1.8e-05$ and $q = 100$.
\item The tree for 7 and 30 days pair: $\zeta  = 8.0e-06$ and $q = 100$.
\item The tree for 14 and 30 days pair: $\zeta =  3.0e-06$ and $q = 900$.
\squishend

\item Causal forest has five hyper-parameters that need to be tuned: (i) $frac$, the fraction of data that is used for training each tree, (ii) $mtry$, the number of variables tried for each split, (iii) $max\_imb$ the maximum allowed imbalance of a split, (iv) $imb\_pen$, a penalty term for imbalanced splits, (v) $q$, the minimum number of observations per condition (control, treatment) in each partition.\footnote{For more information please visit https://github.com/grf-labs/grf/blob/master/REFERENCE.md} 

We used the hyper-parameter optimization procedure available in the grf package for tuning these hyper-parameters. The optimal hyper-parameters for each model are shown below:
\squishlist
\item 7-14 days pair: $frac = 0.5$, $max\_imb= 0.11$, $imb\_pen= 2.03$, $mtry = 13$, and $q=1651$.
\item 7-30 days pair: $frac = 0.5$, $max\_imb= 0.13$, $imb\_pen= 2.49$, $mtry = 1$, and $q=4$.
\item 14-30 days pair: $frac = 0.5$, $max\_imb= 0.20$, $imb\_pen= 5.11$, $mtry = 7$, and $q=121$.
\squishend
\squishend

\newpage 

\section{CATE Estimates from Casual Tree and Causal Forest}
\label{sec:app_cate_est}
\begin{figure}[!htbp]
    \centering
    \includegraphics[scale = 0.27]{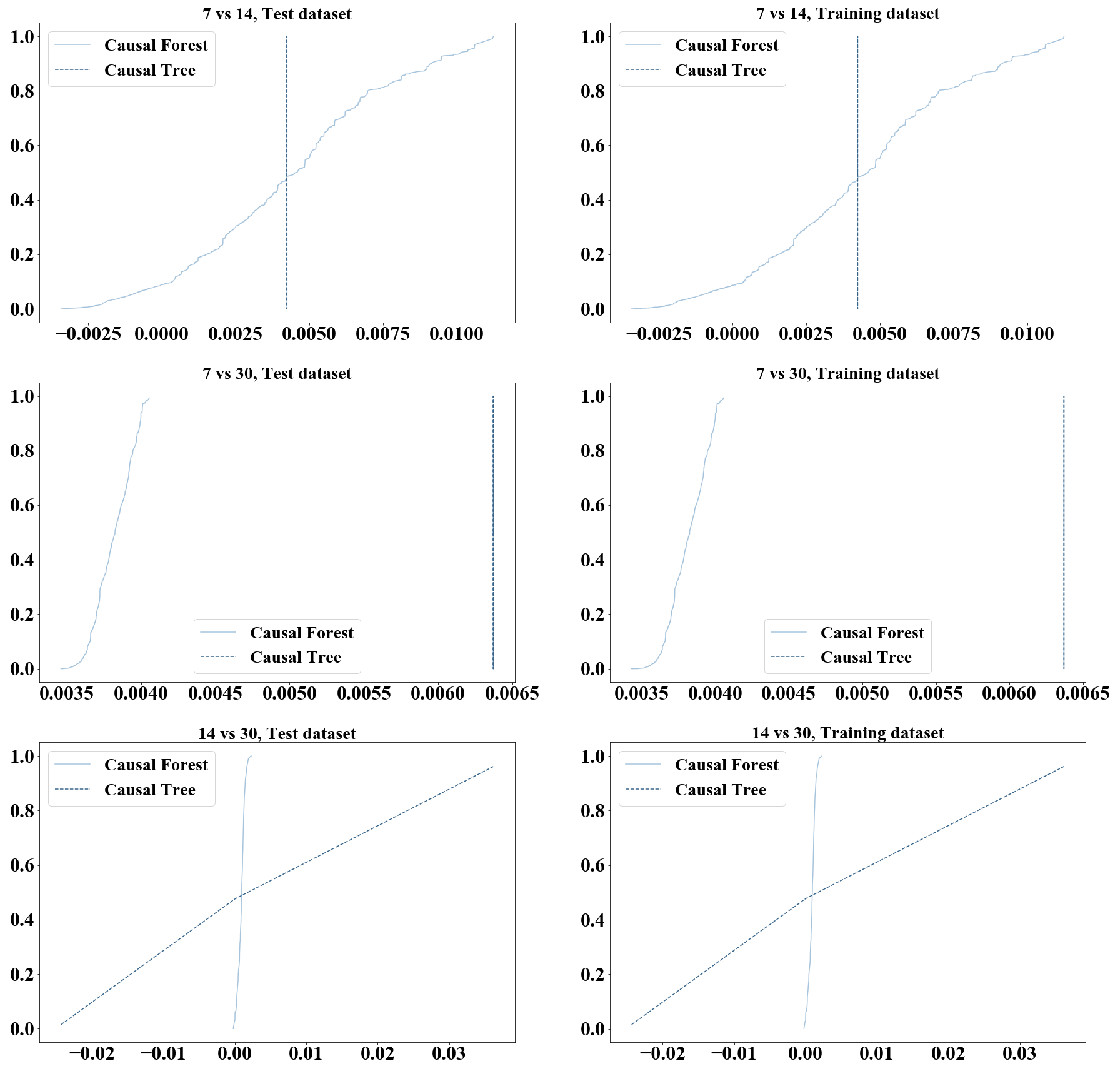}
    \caption{\label{figure:tau_cdfs} The CDFs of treatment effects for each pair of treatments from the causal tree and causal forest models (on training and test data).} 
\end{figure}

\newpage

\section{Paired t-tests Comparing the Rewards under Different Policies}
\label{sec:app_ttest}

\begin{table}[!htbp]
\centering
\footnotesize

\begin{tabular}{ll|cc|cc|cc}
\toprule
\textbf{Policy} & \multirow{2}{*}{\textbf{Policy}} & 
\multicolumn{2}{c}{\textbf{Subscription (\%)}} & 
\multicolumn{2}{c}{\textbf{Subscription Length}}&
\multicolumn{2}{c}{\textbf{Revenue}}\\

\textbf{Category} & & Mean & SD & Mean & SD & Mean & SD \\
\hline\hline
\multirow{3}{*}{\shortstack[l]{One length \\for all}} & $\pi_{30}$ (Baseline) & 14.623 &  0.134 &  2.188 &  0.023 &  78.430 &  1.062 \\
& $\pi_{14}$ & 15.110 &  0.292 &  2.319 &  0.053 &  84.823 &  2.494 \\
& $\pi_{7}$  & 15.441 &  0.287 &  2.324 &  0.053 &  84.437 &  2.499 \\
\midrule
\multirow{4}{*}{\shortstack[l]{Based on \\outcome \\estimation}}& $\pi_{reg}$  & 15.331 &  0.271 &  2.307 &  0.048 &  84.457 &  2.324 \\
& $\pi_{lasso}$  & 15.613 &  0.281 &  2.360 &  0.051 &  87.565 &  2.501 \\
& $\pi_{r\_forest}$  & 14.819 &  0.258 &  2.254 &  0.046 &  80.415 &  2.099 \\
& $\pi_{xgboost}$ & 15.525 &  0.273 &  2.360 &  0.050 &  86.745 &  2.474 \\
\midrule
\multirow{2}{*}{\shortstack[l]{Based on estimates of \\ treatment effect}} & $\pi_{c\_forest}$  & 15.463 &  0.286 &  2.318 &  0.052 &  84.036 &  2.482 \\
& $\pi_{c\_tree}$  & 15.441 &  0.287 &  2.324 &  0.053 &  84.437 &  2.499 \\
\bottomrule
\end{tabular}
\caption{\small\label{table:boostrap_summary} Mean and standard deviation of IPS reward estimations for subscription rate, subscription length (in months), and revenue (scaled) based on 1000 rounds of bootstrap samples drawn from the test data.}
\end{table}

We now examine whether the rewards from the policies are statistically different from each other. A straightforward approach to do this is to run paired Student's t-tests. To do that, we generate 1000 bootstrap samples of our test data and estimate the IPS reward of each counterfactual policy on each sample for the three outcomes of interest -- subscription rate, subscription length, and revenue (scaled). The summary statistics of these IPS rewards are shown in Table \ref{table:boostrap_summary}.

A key assumption necessary for the validity of the Student's t-test is for the data to be generated from a normal distribution. So we use the Shapiro-Wilk test \citep{Shapiro_etal_1965} and Anderson test \citep{Stephens_1974} to test the hypothesis that the distributions of bootstrap IPS reward estimates are normal. Neither of these tests could reject the null hypothesis that the IPS rewards for each of the counterfactual policies come from a normal distribution (even at a $10\%$ confidence interval). 

The results from the paired t-tests for subscription rate are shown in Table \ref{table:paired_t_test_sub} in $\S$\ref{sssec:comp_counter} in the main text of the paper.\footnote{ While paired t-tests are the correct approach here, our results remain the same even if we run unpaired t-tests.}

\begin{table}[!htbp]
\centering
\footnotesize

\begin{tabular}{l|cccc} 
\toprule
Compared & 
\multirow{2}{*}{$\Delta \hat{R} (\pi_{lasso}, \pi)$} &  
\multirow{2}{*}{$\Delta \hat{R} (\pi_{xgboost}, \pi)$} & 
\multirow{2}{*}{$\Delta \hat{R} (\pi_{7}, \pi)$}&
\multirow{2}{*}{$\Delta \hat{R} (\pi_{14}, \pi)$} \\
Policy ($\pi$) &  & & & \\
\hline\hline
$\pi_{xgboost}$ & 0.02 (4.32 $\times$ $10^{-01}$) &                                 --- &                           --- &                                   --- \\
$\pi_7$  & 3.65 (1.14 $\times$ $10^{-89}$) &               3.62 ($\leq$ $10^{-100}$) &                           --- &                                   --- \\

$\pi_{14}$ & 4.11 (4.97 $\times$ $10^{-98}$) &         4.08 (2.94 $\times$ $10^{-79}$) &   0.46 (2.54 $\times$ $10^{-02}$) &                                   --- \\

$\pi_{c\_forest}$ &  4.21 ($\leq$ $10^{-100}$) &               4.18 ($\leq$ $10^{-100}$) &   0.56 (5.35 $\times$ $10^{-13}$) &            0.1 (3.26 $\times$ $10^{-01}$) \\

$\pi_{reg}$ & 5.28 ($\leq$ $10^{-100}$) &               5.25 ($\leq$ $10^{-100}$) &   1.63 (5.84 $\times$ $10^{-24}$) &           1.17 (1.18 $\times$ $10^{-11}$) \\

$\pi_{r\_forest}$ & 10.57 ($\leq$ $10^{-100}$) &              10.55 ($\leq$ $10^{-100}$) &         6.92 ($\leq$ $10^{-100}$) &  6.46 ($\leq$ $10^{-100}$) \\

\bottomrule
\end{tabular}
\caption{\small\label{table:paired_t_test_month} Comparison of reward estimates based on subscription length for the top six policies with other counterfactual policies (on test data). The difference numbers are multiplied by $10^2$. For each cell, we show the mean difference and the $p$-values for the paired t-test in parenthesis. Note that we use $\Delta \hat{R} (\pi_{a}, \pi_{b})$ to show $ \hat{R}_{IPS}(\pi_{a}, Y) - \hat{R}_{IPS}(\pi_{b}, Y)$ in the header.}
\end{table}

\begin{table}[!htbp]
\centering
\footnotesize

\begin{tabular}{l|cccc} 
\toprule
Compared & 
\multirow{2}{*}{$\Delta \hat{R} (\pi_{lasso}, \pi)$} &  
\multirow{2}{*}{$\Delta \hat{R} (\pi_{xgboost}, \pi)$} & 
\multirow{2}{*}{$\Delta \hat{R} (\pi_{14}, \pi)$}&
\multirow{2}{*}{$\Delta \hat{R} (\pi_{reg}, \pi)$} \\ 
Policy ($\pi$) &  & & & \\
\hline\hline
$\pi_{xgboost}$ & 0.82 (1.31 $\times$ $10^{-39}$) &                                 --- &                            --- &                           --- \\
$\pi_{14}$ & 2.74 ($\leq$ $10^{-100}$) &         1.92 (1.22 $\times$ $10^{-85}$) &                            --- &                           --- \\
$\pi_{reg}$  & 3.11 ($\leq$ $10^{-100}$) &               2.29 ($\leq$ $10^{-100}$) &    0.37 (4.19 $\times$ $10^{-06}$) &                           --- \\
$\pi_7$  & 3.13 ($\leq$ $10^{-100}$) & 2.31 ($\leq$ $10^{-100}$) & 0.39 (2.47 $\times$ $10^{-04}$) &   0.02 (3.89 $\times$ $10^{-01}$) \\

$\pi_{c\_forest}$ &  3.53 ($\leq$ $10^{-100}$) &               2.71 ($\leq$ $10^{-100}$) &    0.79 (5.73 $\times$ $10^{-14}$) &    0.42 (3.0 $\times$ $10^{-10}$) \\

$\pi_{r\_forest}$ & 7.15 ($\leq$ $10^{-100}$) & 6.33 ($\leq$ $10^{-100}$) & 4.41 ($\leq$ $10^{-100}$) &         4.04 ($\leq$ $10^{-100}$) \\
\bottomrule
\end{tabular}
\caption{\small\label{table:paired_t_test_rev} Comparison of reward estimates based on revenue (scaled by a fixed number) for the top six policies with other counterfactual policies (on test data). For each cell, we show the mean difference and the $p$-values for the paired t-test in parenthesis. Note that we use $\Delta \hat{R} (\pi_{a}, \pi_{b})$ to show $ \hat{R}_{IPS}(\pi_{a}, Y) - \hat{R}_{IPS}(\pi_{b}, Y)$ in the header.}
\end{table}

\clearpage
\section{Appendix for \texorpdfstring{$\S$}\ref{ssec:segmentation}}
\label{sec:app_segment}

\begin{table}[!htbp]
\centering
\footnotesize
\begin{tabular}{clcccc} 
\toprule
\multirow{2}{*}{\centering \textbf{Variable} }&
\multirow{2}{*}{\centering \textbf{Value} }& \multirow{2}{*}{\centering \textbf{Population} } &  \multicolumn{3}{c}{\textbf{ Policy Assigned Treatment} }\\ 
& &  & \textbf{30 Days} &  \textbf{14 Days} & \textbf{7 Days}\\
\hline\hline
\multirow{5}{*}{\textbf{Country}} &
 \textit{United States} & 54.9\% &  55.9\% &  45.8\% &  57.9\% \\
& \textit{Germany} &8.9\% &   8.2\% &  16.0\% &   6.6\% \\
& \textit{Japan} & 7.9\% &   9.7\% &  11.5\% &   6.4\% \\
& \textit{Other} & 28.3\% & 26.2\% & 26.8\%& 29.1\% \\
\cmidrule{3-6}
& \textit{Total} & 100\% & 100\% & 100\%& 100\% \\
\hline
\multirow{7}{*}{\textbf{Operating System}} &
\textit{Windows 10} & 29.0\% &   9.0\% &  11.9\% &  37.0\% \\
\bottomrule
\end{tabular}
\caption{Distribution of users' pre-treatment attributes for the three segments: those assigned to the 7-day condition, those assigned to the 14-day condition, and those assigned to the 30-day condition. (For each categorical variable, we show the fractions only for the top few categories in the interest of space.)
\label{table:correlate_pre}}
\end{table}
\begin{table}[t]
\centering
\footnotesize
\begin{tabular}{rcccc} 
\toprule
\multirow{2}{*}{\centering \textbf{Variable} }& \multirow{2}{*}{\centering \textbf{Population} } &  \multicolumn{3}{c}{\textbf{ Policy Assigned Treatment} }\\ 
& &  \textbf{30 Days} &  \textbf{14 Days} & \textbf{7 Days}\\
\hline
& \multicolumn{4}{c}{Mean }\\
\cmidrule{2-5}
Subscription rate &14.8\%& 17.0\% & 29.1\%& 9.8\% \\
Revenue &\$536 & \$545 & \$546& \$524 \\
Retention (months) & 16.1 & 17.3 & 16.0& 16.1 \\
Downloaded Products & 1.15 & 1.14 & 1.20 & 1.14 \\
\multicolumn{1}{l}{\textbf{Usage}}\\
\textit{Total used functions} & 2,013 & 2,167 & 3,000 & 1,665\\
\textit{Distinct used functions} & 92& 89 & 108 & 86\\
\hline
& \multicolumn{4}{c}{Fraction}\\
\cmidrule{2-5}
\multicolumn{1}{l}{\textbf{Purchased Bundle}}\\
\textit{Bundle I} & 55.9\% &  57.5\% &  55.4\% &  55.9\% \\
\textit{All inclusive} & 21.5\% &  20.3\% &  21.6\% &  21.7\% \\
\textit{Single Product} & 19.2\% &  16.7\% &  19.2\% &  19.8\% \\
\textit{Other} & 3.4\% &   5.5\% &   3.8\% &   2.6\% \\
\cmidrule{2-5}
\textit{Total} & 100\% & 100\% & 100\%& 100\% \\
\multicolumn{1}{l}{\textbf{Subscription Type}}\\
\textit{Commercial} & 79.1\% &  79.0\% &  81.0\% &  77.3\% \\
\textit{Education} & 20.7\% &  20.6\% &  18.9\% &  22.7\% \\
\textit{Other} & 0.1\% &   0.4\% &   0.1\% &   0.1\% \\
\cmidrule{2-5}
\textit{Total} & 100\% & 100\% & 100\%& 100\% \\
\bottomrule
\end{tabular}
\caption{Means and distributions of users' post-treatment attributes for the three segments: users assigned to the 7-day condition, users assigned to the 14-day condition, and users assigned to the 30-day condition. \label{table:correlate_post}}
\end{table}

\clearpage
\section{Appendix for \texorpdfstring{$\S$}\ref{ssec:revenue}}
\label{sec:app_gain_sub_rev}

\begin{table}[!htbp]
\centering
\footnotesize
\begin{tabular}{cc|ccccccc}
\toprule
\multirow{2}{*}{\textbf{Trial Length}} & \multirow{2}{*}{\textbf{Dataset}} & \multicolumn{7}{c}{\textbf{Subscription Length}}
\\
& & \textbf{Mean} & \textbf{Std} & \textbf{Min} & \textbf{25\%} & \textbf{50\%} & \textbf{75\%} & \textbf{Max} \\
\hline\hline
\multirow{2}{*}{7 Days} & Training & 16.26 & 8.72 & 0 & 10 & 18 & 23 & 73\\
& Test & 16.04 & 8.65 & 0 & 9 & 18 & 23 & 53\\
& All & 16.19 & 8.70 & 0 & 10 & 18 & 23 & 73\\
\hline
\multirow{3}{*}{14 Days} & Training & 15.96 & 8.47 & 0 & 9 & 17 & 22 & 67\\
& Test & 16.39 & 8.48 & 0 & 10 & 18 & 22 & 56\\
& All & 16.09 & 8.48 & 0 & 10 & 18 & 22 & 67\\
\hline
\multirow{3}{*}{30 Days} & Training & 15.89 & 8.31 & 0 & 10 & 17 & 22 & 96    \\
& Test & 16.12 & 8.47 & 0 & 10 & 18 & 22 & 108\\
& All & 15.96 & 8.36 & 0 & 10 & 17 & 22 & 108\\
\bottomrule
\end{tabular}
\caption{\label{table:retention_summary} The summary statistics of the subscribed users' total months of subscription.}
\end{table}
\begin{table}[!htbp]
\centering
\footnotesize
\begin{tabular}{c|ccccc|ccc}
\toprule
\textbf{Trial}& \multicolumn{5}{c}{\textbf{Subscribed Bundle}} & \multicolumn{3}{c}{\textbf{Subscription Type}}
\\
\textbf{Length} & \textbf{1} & \textbf{All Inclusive} & \textbf{Single Product} & \textbf{4} & \textbf{5} & \textbf{Commercial} & \textbf{Education} & \textbf{Government}\\
\hline\hline
7 Days &  55.24 & 22.02 & 19.44 & 1.68 & 1.62 & 78.57 & 21.30 & 0.13\\
14 Days & 55.99 & 21.76 & 19.12 & 1.79 & 1.34 & 79.62 & 20.25 & 0.13\\
30 Days & 55.98 & 21.64 & 18.95 & 1.82 & 1.62 & 79.02 & 20.85 & 0.13\\
\bottomrule
\end{tabular}
\caption{\label{table:subscription_summary} The fraction of each subscription bundle and type. We do not reveal the names of some of the bundles to preserve's the firm's anonymity.}
\end{table}

\end{appendices}
\end{document}